# Fusing Knowledge and Language: A Comparative Study of Knowledge Graph-Based Question Answering with LLMs

Vaibhav Chaudhary[a], Neha Soni[b], Narotam Singh[c#], Amita Kapoor[d#]

[a]*Department of Mechanical Engineering, Indian Institute of Technology, Kharagpur, India*
[b]*Department of Physics, MMEC, Maharishi Markandeshwar (Deemed to be University), Mullana-Ambala, Haryana, India*
[c]*Former Scientist, IMD, Ministry of Earth Sciences, Delhi, India. Email: narotam.singh@gmail.com*
[d]*NePeur, India, Email: amita.kapoor@nepeur.com*
[#]*Corresponding authors*

## Abstract

Knowledge graphs, a powerful tool for structuring information through relational triplets, have recently become the new front-runner in enhancing question-answering systems. While traditional Retrieval Augmented Generation (RAG) approaches are proficient in fact-based and local context-based extraction from concise texts, they encounter limitations when addressing the thematic and holistic understanding of complex, extensive texts, requiring a deeper analysis of both text and context. This paper presents a comprehensive technical comparative study of three different methodologies for constructing knowledge graph triplets and integrating them with Large Language Models (LLMs) for question answering: spaCy, Stanford CoreNLP-OpenIE, and GraphRAG, all leveraging open source technologies. We evaluate the effectiveness, feasibility, and adaptability of these methods by analyzing their capabilities, state of development, and their impact on the performance of LLM-based question answering. Experimental results indicate that while OpenIE provides the most comprehensive coverage of triplets, GraphRAG demonstrates superior reasoning abilities among the three. We conclude with a discussion on the strengths and limitations of each method and provide insights into future directions for improving knowledge graph-based question answering.

## 1. Introduction

Retrieval-Augmented Generation (RAG) has emerged as a powerful approach in natural language processing, particularly in question-answering systems. By combining the retrieval of relevant documents with the generative capabilities of large language models (LLMs), RAG models are able to generate contextually accurate and informative responses. Traditional RAG approaches typically rely on retrieving passages from vast corpora of unstructured text, which are then fed into a generative model to formulate an answer. While effective, this approach often faces challenges such as irrelevant retrieval, redundancy in answers, and limited capacity to understand structured relationships between concepts.

Knowledge graph based question-answering (KG-QA) systems have gained significant attention in addressing these challenges. Unlike traditional RAG, which depends heavily on unstructured text retrieval, KG-QA leverages structured knowledge in the form of graph triplets, representing relationships between entities. The system can provide more precise, explainable, and contextually relevant answers by querying these triplets. Knowledge graphs offer a way to formalize the relationships between entities, making tracing and verifying the rationale behind a generated answer easier. When combined with LLMs, KG-QA systems can take advantage of both the structured reasoning capabilities of knowledge graphs and the natural language generation strengths of the LLMs.

Creating high-quality knowledge graph triplets is crucial for the success of such systems. This process can be achieved through various methods, each with different strengths. Rule-based systems like *spaCy* leverage pre-defined linguistic rules and patterns to offer precision in extracting triplets from domain-specific texts. These systems could perform well and be finely tuned to some specific domain, such as the medical or legal domain, where the structure and vocabulary could be well defined. However, the actual system performance would depend on how well the rules are defined and how much knowledge the models can capture and embed in itself. At the same time, open-domain extractors like Stanford *CoreNLP's OpenIE* are more flexible, capturing a wider range of relations. They utilize language semantics and structure to push for identifying a subject, object, and relation in any sentence using limited or no training data. More recently, *GraphRAG* by Microsoft has emerged as a technique that leverages the power of LLMs for extracting precise structural information hierarchically from unstructured data by utilizing LLM inference at various stages of knowledge graph creation. It can create a more holistic view of the text and answer questions at different levels of abstraction, which can prove quite helpful more often than not.

A key motivation for this research is to overcome the inherent limitations of traditional RAG approaches, which are adept at extracting localized, fact-based information yet often fall short when tasked with understanding the broader thematic context or handling complex, large-scale texts. By integrating knowledge graphs—capable of capturing structured relationships—into QA systems, we aim to enhance LLMs' reasoning capabilities, resulting in accurate, contextually richer responses, and more explainable. Furthermore, the comparative evaluation of diverse extraction methods such as spaCy, CoreNLP OpenIE, and GraphRAG is crucial to identifying the most effective techniques for constructing high-quality knowledge graphs. Leveraging open-source technologies for this study ensures replicability and practical applicability, ultimately guiding researchers and practitioners toward improved KG-QA systems better equipped to address real-world challenges. Therefore, this study not only fills a current research gap but lays the groundwork for future knowledge graph-based question-answering innovations.

In this paper, therefore, we present a comparative study of three popular methods for generating graph triplets and integrating them with LLMs for question answering: (1) spaCy, (2) CoreNLP OpenIE, and (3) GraphRAG. Each method represents a unique approach to creating knowledge graphs and using them in QA systems. By evaluating them across broad parameters—such as performance, ease of use, accuracy, customizability, etc.—we aim to provide insights into their strengths and weaknesses and how they contribute to the effectiveness of KG-QA systems. This study aims to guide researchers and practitioners in selecting the most appropriate tools for their knowledge graph-based question-answering tasks.

The key points of contribution of our work are as follows:
1. We present a fully reproducible framework that extracts relational triplets using spaCy, Stanford CoreNLP-OpenIE, and GraphRAG, constructs knowledge graphs, and integrates them into LLM prompts for enhanced question answering.
2. Under identical conditions, we systematically evaluate each method on a whole range of parameters, filling a critical gap in the literature. Our evaluations are conducted on two distinct data sources—Shakespeare's play "All's Well That Ends Well" and the RepliQA dataset—to ensure robustness and generalizability of our findings.
3. By embedding structured relations into QA, we demonstrate that GraphRAG performs the best in complex, thematic queries, CoreNLP OpenIE offers the broadest factual coverage, and spaCy provides a lightweight, high-precision baseline.
4. In addition to GPT-4 automated scoring, we incorporate domain expert judgments to assess answer coherence, reasoning ability, and contextual richness, yielding a more nuanced understanding of KG-augmented QA performance.
5. Based on our dual-dataset empirical analysis, we recommend optimal extraction techniques—balancing factual coverage, reasoning depth, and computational efficiency—along with best practices for graph-based question answering.
6. We identify promising directions for advancing KG-based QA, like dynamic graph updates, hybrid extraction pipelines, multi-modal knowledge integration, and domain-specific adaptations.

## 2. Background or Related Work

*Introduction to KGs and QA systems*
*Early QA systems*
Question-answering (QA) systems have long been a cornerstone of natural language processing (NLP) research. The earliest QA systems cited in the literature are BASEBALL (1961) [1] and LUNAR (1972) [2]. BASEBALL answered questions about baseball games over a one-year period, while LUNAR provided access to data from Apollo moon mission rock sample analyses. Both used predefined natural language patterns to translate questions into structured database queries and are considered the first examples of Natural Language Interface to Databases (NLIDB) [3]. These initial systems relied on well-defined ontologies, leveraging formal logic and semantic parsing to match user queries with predefined answers.

In the 1980s and 1990s, knowledge base systems, often built for specific domains, gained popularity. The systems were well-suited for question answering, where users presented a problem and received answers through menus or natural language interfaces. The system interacts with users, asking additional questions to clarify intent and reasons using the knowledge base and user input. This deductive question-answering approach emphasizes reasoning and explanation and is rooted in early expert systems like MYCIN [4] (medical reasoning) and SHRDLU [5] (robot interaction with toy blocks). Though effective for specific domains, these systems were limited by their rigid structure and lack of adaptability to varied and open-ended questions.

The innovative question-answering era began in 1999 with TREC's introduction of open-domain question answering [47]. Participants aimed to answer natural language questions concisely using large text collections like AQUAINT and AQUAINT2. The questions spanned various topics and often required shallow text analysis. TREC had a major impact on the field, influencing evaluation metrics and leading to the development of the first web-based QA system, START, by MIT [48]. In 2002, the INEX(INitiative for the Evaluation of XML retrieval) introduced tasks leveraging document structure for improved information retrieval.

A significant milestone in the development of QA systems was the introduction of Knowledge Graphs (KGs), a term coined by Google. KGs represent information as a network of entities (nodes) and relationships (edges), making it possible to answer questions by reasoning over these structured triplets. Knowledge graphs have been built through different methods, including manual curation (e.g., Cyc [49]), crowd-sourcing (e.g., Freebase [50], Wikidata [51]), and extraction from semi-structured sources like Wikipedia (e.g., DBpedia [52], YAGO [53]). Information extraction techniques from unstructured data have also produced graphs like NELL [54], PROSPERA [55], and KnowledgeVault [56]. All of these earlier graphs are mostly Encyclopedic KGs[12]. Apart from these, as described in [12], other types of KGs, such as Commonsense KGs (ex., ConceptNet[13]), Domain-specific KGs (ex., UMLS[14]), and multi-modal KGs (ex., IMGpedia[15] and MMKG[16]), have also appeared. These systems have demonstrated the power of knowledge graphs in expanding the domain coverage of QA systems by providing a more flexible and scalable knowledge representation. These systems laid the groundwork for large-scale information retrieval based on semantic understanding.

*Machine learning in QA systems*
However, while powerful in factual question answering, traditional knowledge graph-based QA systems struggled with the nuances and variability of natural language. This limitation paved the way for the integration of machine learning into QA systems, particularly the use of neural networks to improve language understanding. A landmark achievement in this area was IBM's Watson, which combined deep learning and statistical techniques to defeat human champions in the game show *Jeopardy!*[6] by answering complex, context-heavy questions. Watson marked the shift from rule-based to machine learning-based QA, where systems began to rely on probabilistic models to interpret and answer questions from unstructured text. Several Deep Neural Network models have been applied to NLP tasks, mostly using RNNs like LSTMs or GRUs for classification or summarization[7]. Other works are Kapashi et al. [9], which proposes a baseline LSTM model; Memory Networks introduced by Weston et al. [10], which incorporated memory structures to answer questions effectively, and the Dynamic Memory Networks model [11], which integrates memory networks with an attention mechanism. In addition to the NLP and deep learning techniques used in question-answering tasks, Mikolov et al.'s work on the word2vec and Glove embedding models has been particularly influential. They use an approach to effectively represent words as word vectors, capturing semantic relationships and contextual information.

*LLMs in QA*
The advent of LLMs has steered in a transformative era in NLP, particularly within the domain of QA. These models, pre-trained on massive corpora of diverse text, exhibit sophisticated capabilities in both natural language understanding and generation. Their proficiency in producing coherent, contextually relevant, and human-like responses to a broad spectrum of prompts makes them exceptionally well-suited for QA tasks, where delivering precise and informative answers is paramount. Recent advancements by models such as BERT [57] and ChatGPT [58], have significantly propelled the field forward. LLMs have demonstrated strong performance in open-domain QA scenarios—such as commonsense reasoning[20]—owing to their extensive embedded knowledge of the world. Moreover, their ability to comprehend and articulate responses to abstract or contextually nuanced queries and reasoning tasks [22] underscores their utility in addressing complex QA challenges that require deep semantic understanding. Despite their strengths, LLMs also pose challenges: they can exhibit contextual ambiguity or overconfidence in their outputs ("hallucinations")[21], and their substantial computational and memory requirements complicate deployment in resource-constrained environments.

*RAG, fine tuning in QA*

LLMs also face problems when it comes to domain specific QA or tasks where they are needed to recall factual information accurately instead of just probabilistically generating whatever comes next. Research has also explored different prompting techniques, like chain-of-thought prompting[24], and sampling based methods[23] to reduce hallucinations. Contemporary research increasingly explores strategies such as fine-tuning and retrieval augmentation to enhance LLM-based QA systems. Fine-tuning on domain-specific corpora (e.g., BioBERT for biomedical text [17], SciBERT for scientific text [18]) has been shown to sharpen model focus, reducing irrelevant or generic responses in specialized settings such as medical or legal QA. Retrieval-augmented architectures such as RAG [19] combine LLMs with external knowledge bases, to try to further mitigate issues of factual inaccuracy and enable real-time incorporation of new information. Building on RAG's ability to bridge parametric and non-parametric knowledge, many modern QA pipelines introduce a lightweight re-ranking step [25] to sift through the retrieved contexts and promote passages that are most relevant to the query. However, RAG still faces several challenges. One key issue lies in the retrieval step itself—if the retriever fails to fetch relevant documents, the generator is left to hallucinate or provide incomplete answers. Moreover, integrating noisy or loosely relevant contexts can degrade response quality rather than enhance it, especially in high-stakes domains where precision is critical. RAG pipelines are also sensitive to the quality and domain alignment of the underlying knowledge base, and they often require extensive tuning to balance recall and precision effectively.

*RAG + Graphs in QA*
To address these limitations of traditional RAG, particularly in retrieving contextually precise and semantically rich information, research has tried to integrate structured knowledge sources and graph-based reasoning into the retrieval pipeline. One notable direction of work is LLM-based retrieval, which incorporates knowledge graphs information into the generation process of LLMs. The LLMs are augmented using the retrieved facts from the KG [26], leading to a clear dependence on the quality of extracted graph information, using it to generate responses. Research has also been done towards augmenting knowledge graph information, retrieved through some semantic or some other similarity index, in the prompts given to the LLM [27], to help the model do zero-shot question-answering. Some researchers have tried a different approach to fact retrieval, where the model tries different queries, using structured query languages, until the desired information comes through [28]. All of these approaches have used KGs as an external source to retrieve information from and answer the questions [29]. Then, aligning the retrieval process with LLM even more closely, some researchers have proposed methods which use LLMs in intermediate steps as well to plan the retrieval and judge whether the retrieved information is relevant or not [29,30], continuing the process until the desired output emerges.

Another interesting direction of work is integrating GNNs with LLMs, which leverages graph neural networks to enhance retrieval and re-ranking using learned graph representations, along with generation capabilities of LLMs. There have been approaches such as GNN-RAG [31], which have tried combining language understanding abilities of LLMs with the reasoning abilities of GNNs in a retrieval-augmented generation (RAG) style. Other methods of GNN-LLM alignment, have been classified into symmetric and asymmetric alignment [32]. Symmetric alignment refers to the equal treatment of the graph and text modalities during the alignment process[33, 34, 35]. Asymmetric alignment focuses on allowing one modality to assist or enhance the other, here leveraging the capabilities of GNNs to improve the LLMs [36, 37, 38].

*Evaluation of QA*

The continual evolution of such hybrid architectures reflects the dynamic nature of the QA landscape and its responsiveness to complex information needs. Given the diversity of methods explored to enhance QA systems—from rule-based techniques to advanced neural and hybrid models—it becomes essential to establish robust mechanisms for evaluating their effectiveness. It is thus important to discuss the evaluation paradigms that compare and benchmark these systems. Extractive QA benchmarks almost universally use Exact Match (EM)—the strict proportion of predictions that character-for-character match a ground-truth answer—and F1 score, the harmonic mean of token-level precision and recall, to evaluate answer quality. On SQuAD v2.0 [39], human annotators achieve around 86.831 EM and 89.452 F1 on the test set, whereas state-of-the-art models now exceed 90 EM and 93 F1. Natural Questions [40] extends this paradigm to long-answer (paragraph) and short-answer (span or yes/no) annotations drawn from real Google search queries. Meanwhile, multiple-choice datasets like OpenBookQA [41]—designed to mimic open-book science exams—use simple accuracy. Complex reasoning benchmarks push beyond single-span extraction. HotpotQA [42], a multi-hop dataset built on pairs of Wikipedia articles, evaluates both answer-span EM/F1 and supporting-fact EM/F1, plus a joint metric requiring both to be correct; even top models achieve only ~72 joint F1[43] under the full-wiki setting, far below human performance of ~82 average F1 and ~96 upper bound F1. These core metrics and datasets underpin broader QA evaluation: multi-task suites like GLUE/MMLU [44, 45] include QA subtasks to probe general language understanding, while specialized frameworks such as MT-bench [46] ("LLM as judge") and automated platforms like Scale Evaluation layer on top to assess conversational and retrieval-augmented QA in real-world scenarios.

This review has highlighted the broad spectrum of methodologies and evaluation strategies used in enhancing and assessing question answering systems. With this foundation in place, we now transition to the specific tools and frameworks employed in our research for graph-based question answering. In the following section, we briefly introduce and contextualize spaCy, Stanford CoreNLP, and GraphRAG—three diverse and widely-used tools we have utilized for knowledge graph construction and integration with LLMs.

*Spacy*
spaCy is a cutting-edge, open-source library for Natural Language Processing in Python that combines efficiency with a user-friendly design. Engineered for both research and production, it offers robust pre-trained models for tasks like tokenization, dependency parsing, and named entity recognition, making it an indispensable tool for rapidly prototyping and deploying NLP solutions.

*CoreNLP*
Stanford CoreNLP is a comprehensive natural language processing toolkit developed by the Stanford NLP Group that offers a wide range of linguistic analysis tools, including tokenization, part-of-speech tagging, named entity recognition, dependency parsing, and sentiment analysis. Engineered with academic research and industrial applications in mind, it provides reliable and accurate text annotations that facilitate deeper understanding and advanced analysis of language data.

*GraphRAG*
GraphRAG, a Microsoft Research innovation, represents a significant advancement in retrieval-augmented generation (RAG) by integrating structured knowledge graphs to enhance large language models' (LLMs) contextual understanding of private datasets. Unlike traditional RAG systems that rely on vector similarity searches, GraphRAG constructs an LLM-generated entity-relationship graph from source documents, enabling holistic analysis and semantic connections across disparate data points. GraphRAG addresses critical limitations in baseline RAG systems by transforming unstructured text into structured knowledge representations, particularly in synthesizing cross-document insights and supporting verifiable assertions through source-linked evidence. It extracts relevant data from the knowledge graph, identifies the appropriate community related to the question, and utilizes the generated graph, entities, relations, claims, summaries, etc., to generate accurate and concise responses.

*Hence, the comparison and prospects of improvements.*
Thereby concluding the literature review section, we will now leverage these three tools—spaCy, Stanford CoreNLP, and GraphRAG—for constructing and utilizing knowledge graphs within question answering systems. By evaluating them across a broad set of parameters—including performance, ease of use, accuracy, customizability, and more—we aim to present a comprehensive comparative analysis. This will help highlight each tool's unique strengths and limitations, and their overall impact on the performance and robustness of knowledge graph-based QA systems.

The subsequent sections of this paper are structured as follows: Section 3 outlines the methodology adopted for constructing knowledge graphs and integrating them with LLMs; Section 4 presents our results along with a detailed discussion; and Section 5 concludes the paper with key takeaways and directions for future research.

## 3. Methodology

This section outlines the methodology used to compare three distinct approaches for knowledge graph-based question answering (KG-QA) using large language models (LLMs): spaCy, CoreNLP OpenIE, and GraphRAG. Each tool is evaluated based on its ability to generate graph triplets and integrate these structured knowledge representations with an LLM to provide meaningful answers to natural language questions. The comparison is conducted across several key parameters, including ease of use, performance, accuracy, customizability, etc. By systematically evaluating these methods, we aim to uncover the strengths and limitations of each approach in the context of KG-QA.

### 3.1. Experimental Setup

The same data sources and LLMs are employed across all three tools to ensure a fair comparison. The data consists of Shakespeare's play "All's Well That Ends Well" and the RepliQA dataset, which provides a diverse set of long-form questions and reference answers for factual consistency evaluation. Questions are crafted to assess the three approaches in depth and to highlight their respective strengths and limitations. All LLMs used for answering are open-source-based, enhancing the accessibility and reproducibility of the methods. Additionally, a comparative analysis is performed to observe model behavior when the evaluation criteria are known versus when they are not disclosed. The responses are assessed using a consistent rubric evaluating content, organization, style, and mechanics. GPT-4 is employed as an expert evaluator to rate the answers based on this rubric.

### 3.2. Data Sources

#### 3.2.1. All's Well That Ends Well

Shakespeare's play "All's Well That Ends Well" is taken from the internet, which is then subjected to different approaches to make a knowledge graph and answer the relevant questions. However, due to resource constraints, only the initial part of the play's text was utilized.

### 3.2.2. RepliQA

The RepliQA dataset, a benchmark designed for evaluating the factual consistency and grounding of long-form question answering systems, is used in our experiments. It comprises questions and reference answers curated from diverse domains, along with multiple generated responses. This dataset allows for rigorous testing of the methods, and also to verify the consistency in experiment results across data sources. For the purpose of this paper, the 'repliqa_0' variant of the dataset was used. For the purpose of this paper, ten documents were randomly chosen with the document size in the 3000-4000 characters range. This was done to avoid the resource constraints.

### 3.3. Workflow for Question Answering using spaCy triplets

spaCy's Named Entity Recognition (NER) and dependency parsing capabilities were used to extract entities and relationships from the text. The entities were found by analyzing dependency tags for subjects and objects. Dependency patterns were defined and used to determine the relationship. On similar lines, triplets were also extracted from the question. Then, the question triplets were flattened and processed and finally used to filter out the relevant triplets from all the extracted ones. The relevant triplets were then used to create a graph, which was given alongside a question, and the LLM to be used, which was used by the Langchain GraphQAChain to generate the answer. The workflow involves minimal configuration and relies heavily on spaCy's built-in models for extracting entities.

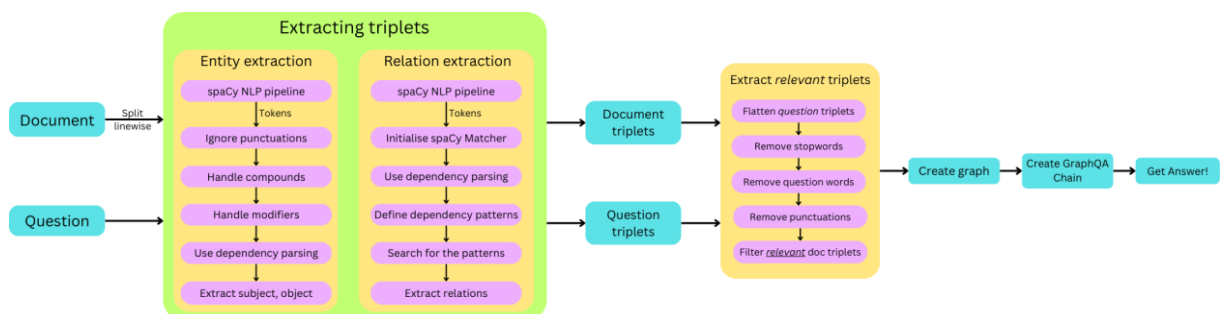

Figure 1: spaCy workflow

### 3.4. Workflow for Question Answering using CoreNLP triplets

Stanford CoreNLP's Open Information Extraction (OpenIE) module was used to identify triplets from the text. The CoreNLP Java server was locally hosted. The processing pipeline starts with the tokenization of the text, does PoS tagging on the tokens, and lemmatizes them, followed by dependency parsing and natural language semantic analysis. The OpenIE triplet extraction module then utilizes the output to finally give out the formed triplets. The question also goes through dependency parsing, extracting phrases, processing them, and finally, relevant document triplets are filtered out for use. The relevant triplets were then used to create a graph, which was given alongside a question, and the LLM to be used, which was used by the Langchain GraphQAChain to generate the answer. CoreNLP offers flexibility in the level of granularity for triplet extraction, making it highly customizable for different tasks.

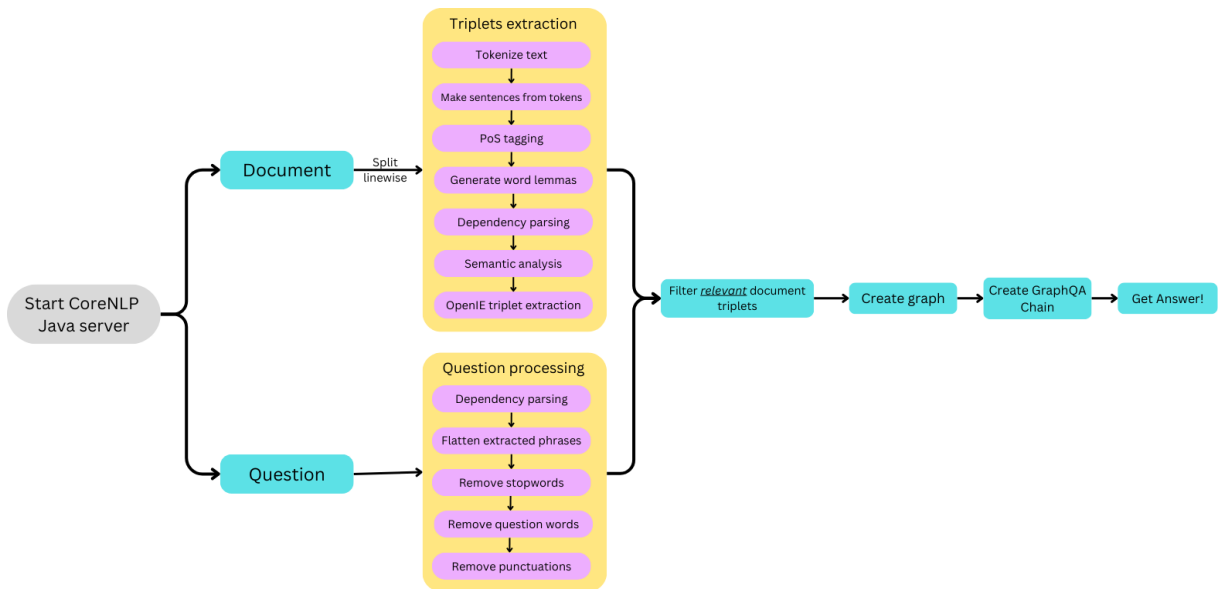

Figure 2: CoreNLP workflow

## 3.5. Workflow for Question Answering with GraphRAG

Microsoft's GraphRAG is a structured method to extract a knowledge graph, organize information, and generate summaries. The document is given to GraphRAG as a whole. The GraphRAG framework systematically transforms unstructured text into hierarchical knowledge representations through four stages: (1) entity-relationship extraction using LLMs to build consolidated knowledge graphs, (2) community detection for multi-level semantic clustering, (3) automated summarization of communities into executive reports and hierarchical context maps, and (4) graph-aware retrieval combining community-level context with dynamic query routing. GraphRAG can utilize both local neighborhood expansions and global hierarchy navigation for multi-hop reasoning. Finally, the local response pipeline was used to get answers to the questions, as the questions posed mostly revolved around the characters in the play, and local response pipeline was the most suitable one for that.

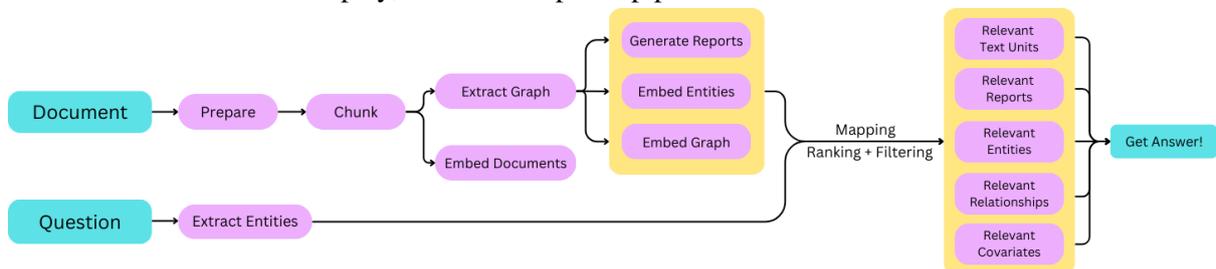

Figure 3: GraphRAG workflow

## 3.6. Evaluation Criteria

The approaches are evaluated across multiple dimensions and compared across several parameters.

### 3.6.1. The Technical Comparative Study

A comprehensive technical study was conducted across multiple parameters to compare spaCy, Stanford CoreNLP, and GraphRAG as distinct approaches for constructing knowledge graphs. The following are the broad parameters that are decided for the purpose:

- Ease of Use and Setup: evaluating the simplicity of installation and configuration

- Diverse Support: assessing the diversity of supported languages and community contributions
- Customizability: flexibility to adapt each tool to specific use cases
- Advanced Features: Ontology integration and reasoning capabilities
- Ease of Integration: each tool's capabilities and compatibility with existing systems
- Hardware Requirements: understand the computational demands of each approach
- Explainability & Transparency: evaluated for clarity in output and model behavior
- Licensing and Cost: insight into each tool's accessibility and expenses

### 3.6.2. The Performance-Based Comparative Study

Next, the approaches are compared based on some performance-related parameters on the evaluation tasks decided:

- Evaluations
- Preprocessing Overload

## 4. Results and Discussion

### 4.1. The Technical Comparative Study

#### 4.1.1. Ease of Use and Setup

(1) *Installation*, which assesses deployment simplicity

SpaCy offers a straightforward installation, requiring only the library installation via PyPi and an optional Visual Studio Code extension for enhanced development support, reflecting its user-centric design. In contrast, CoreNLP presents a more complex setup process involving multiple steps, such as downloading the package, unzipping files, installing Java, downloading a language model, running a Java server to host the API, and sending POST requests with the specified pipeline. GraphRAG lies in between, requiring installation as a package from PyPi or GitHub, followed by workspace initialization within a directory via the command line. Pipelines and question-answering functionalities can be executed directly from the command line, with any configuration changes necessitating reruns for updated outputs.

Table 1: Comparison table for the 'Installation' subparameter

| Parameter | spaCy | CoreNLP | GraphRAG |
|---|---|---|---|
| **Installation Method** | pip install spacy | Download + Java Setup | pip install graphrag |
| **Dependencies** | None | Java and language models | Directory initialization |
| **Additional Steps** | Optional: VS code extension | Run Java server | CLI commands for pipeline |
| **Configuration Updates** | Command re-run | Manual restart | CLI re-run |

(2) Documentation, which evaluates the clarity and comprehensiveness of resources

spaCy excels with its highly descriptive and frequently updated resources, reflecting the continuous support and maintenance from its developer, Explosion AI. In comparison, CoreNLP offers descriptive documentation as well, but its updates are less frequent, owing to its maintenance by an academic institution rather than a commercial entity. GraphRAG, while featuring comprehensive and active documentation, remains in its early stages of development, with basic-level resources that require significant enhancements to achieve parity with more mature tools.

Table 2: Comparison table for the 'Documentation' subparameter

| Parameter | spaCy | CoreNLP | GraphRAG |
|---|---|---|---|
| Documentation Quality | Highly descriptive, user-friendly | Comprehensive, technical | Comprehensive but basic-level |
| Frequency of Updates | Frequent (monthly/quarterly) | Infrequent (yearly or less) | Frequent (early-stage development) |
| Maintenance Entity | Explosion AI | Academic institution (Stanford NLP Group) | Microsoft |

(3) *Learning Curve*, measuring the time users need to become proficient.

The learning curve varies significantly, reflecting differing design philosophies and use cases for each tool. spaCy offers a smooth and beginner-friendly learning experience, supported by its extensive documentation and readily available resources, making it intuitive to start and progress. Conversely, CoreNLP presents a slightly steeper learning curve, primarily due to its more complex initial setup involving server configurations. While the process is straightforward, it can appear more daunting than spaCy's simplicity; however, once the setup is complete, CoreNLP proves efficient and user-friendly for further use. GraphRAG, on the other hand, offers an accessible starting point with the help of its documentation and tutorials. However, delving deeper into its capabilities demands a strong foundational understanding of LLMs, graph structures, and related concepts, making it better suited for users with prior technical expertise.

Table 3: Comparison table for the 'Learning Curve' subparameter

| Parameter | spaCy | CoreNLP | GraphRAG |
|---|---|---|---|
| Learning Curve | Smooth, beginner-friendly | Moderate | Easy to start but steep for advanced usage |
| Required Expertise | Minimal (basic Python knowledge) | Moderate (Java server configurations) | Basic (direct usage) *or* High (customizing per use-case) |

### 4.1.2. Diverse Support

(1) Languages Supported

spaCy claims support for over 70 languages, with pre-trained models and pipelines readily available for 24, alongside a multi-language pipeline designed for broader linguistic flexibility. CoreNLP, in contrast, supports 8 languages as of its latest version, indicating a narrower but still diverse linguistic coverage. GraphRAG, being in its nascent stages, currently supports English natively; however, it allows for integrating other languages through language-specific prompts and models while maintaining its pipeline and workflow structure.

Table 4: Comparison table for the 'Languages Supported' subparameter

| Parameter | spaCy | CoreNLP | GraphRAG |
|---|---|---|---|
| **Supported Languages** | 70+ supported, 24 pre-trained | 8 supported | English natively, others possible |
| **Pre-Trained Models** | Yes | Yes | No |
| **Multi-Language Support** | Pipeline available | Limited | Prompt/ model-based |
| **Additional Notes** | Broadest coverage among the three tools | Focused primarily on core linguistic features | Customizable via LLM prompts for non-English use |

(2) Community Support and Contributions

spaCy is by far the most "popular" of the three—with a large, highly active GitHub community, extensive Stack Overflow support, and a vibrant ecosystem of community contributions and extensions. Its Python-only design and focus on production use have made it the go-to choice for many developers. Stanford CoreNLP—written in Java and long established in academia—has a more modest but steady community. Its support channels tend to be more traditional (mailing lists, academic forums) and its GitHub metrics are lower than spaCy's, reflecting its more specialized user base. GraphRAG is an emerging tool that integrates graph-based retrieval with LLMs. Although it shows promise, its GitHub numbers, Stack Overflow activity, and community contributions are currently much lower than those of spaCy or CoreNLP. It remains more niche and is still growing its ecosystem.

Table 5: Comparison table for the 'Community Support and Contributions' subparameter

| Parameter | spaCy | CoreNLP | GraphRAG |
|---|---|---|---|
| **GitHub Stars** | ~29,000+ | ~1,500–2,500 | A few hundred |
| **GitHub Forks** | ~4,000+ | A few hundred | Under 100 |
| **GitHub Pull Requests** | Active, high contribution volume | Lower contribution rate | Relatively low |
| **GitHub Issues** | Hundreds (open & closed) | Moderate number | Limited |
| **Stack Overflow Questions** | 3,000+ | Few hundred | Very few |
| **Community Contributions** | Extensive third-party tools & integrations | Primarily academic contributions | Emerging ecosystem, limited third-party tools |
| **Discord/Forums** | Active discussions (GitHub, Discord, Explosion AI forums) | Primarily mailing lists & forums | Limited, mostly GitHub issues & Microsoft forums |

### 4.1.3. Customizability

spaCy provides extensive support for training and customizing pipelines, allowing users to tailor models for specific text types with a streamlined approach. Its flexible configuration system enables easy integration of custom components, with HashEmbedCNN as the current standard. Additionally, users can design new evaluation metrics and specific components to enhance performance on domain-specific tasks. CoreNLP also supports training and integrating custom models by adjusting configurations and specifying custom files, but its complexity makes the process more resource-intensive. While it allows modifications to pipeline steps, adding entirely new components is tedious. In contrast, GraphRAG offers a highly flexible framework, allowing users to integrate any custom LLMs, prompts, and configurations. It facilitates the creation of bespoke workflows, enabling domain-specific pre-processing and post-processing steps with minimal constraints.

Table 6: Comparison table for the 'Customizability' subparameter

| Feature | spaCy | CoreNLP | GraphRAG |
|---|---|---|---|
| **Custom Pipeline Support** | Robust support with configurable pipelines | Custom models through configuration and files | Custom LLMs, prompts, and configurations |
| **Ease of Custom Model Training** | Easy to train with built-in commands and flexible config files | Requires significant technical effort and manual configuration | Fully supports fine-tuned LLMs with prompt engineering |
| **Configuration Complexity** | Uses a straightforward *.cfg* file for pipeline setup | Requires Java-based configurations | Highly flexible with YAML/JSON-based workflow setups |
| **Pre-trained Model Availability** | Large number of pre-trained models available, including transformer-based models | Several models available, but less diverse than spaCy | Supports any LLM with pre-trained embeddings and vector stores |
| **Execution Time for Customization** | Fast – optimized for quick pipeline modifications | Slower – requires substantial manual setup | Moderate – depends on external LLM config |

### 4.1.4. Advanced Features

(1) Ontology Integration

Ontology integration is vital because it transforms isolated data points into interconnected, hierarchical structures that enable more sophisticated inference, improved domain-specific reasoning, and enhanced contextual understanding. GraphRAG excels in this regard by automatically forming communities and generating summaries for each, effectively embedding domain ontologies within its workflow. This integration not only streamlines the process of identifying relationships among entities but also provides a structured foundation for downstream tasks such as reasoning and summarization.

In contrast, SpaCy and CoreNLP are primarily designed for extracting entities and relationships without an inherent mechanism for constructing larger, ontology-driven groupings. However, these tools could be enhanced to support ontology integration through the following approach:

**Post-Processing Pipeline:**

1. *Entity Extraction:* Use SpaCy or CoreNLP to identify and extract entities from text.
2. *Entity Mapping:* Map the extracted entities to nodes within a domain-specific ontology using similarity measures or knowledge-base lookups.
3. *Clustering and Hierarchical Structuring:* Apply clustering algorithms (e.g., community detection or hierarchical clustering) to group related entities and infer hierarchical relationships.
4. *Summarization and Inference:* Integrate LLMs to generate summaries for each cluster, thereby emulating the end-to-end workflow seen in GraphRAG.

By combining the strong extraction capabilities of SpaCy/CoreNLP with graph-based algorithms and LLM-powered summarization, researchers can create hybrid systems. Such systems would leverage the speed and flexibility of traditional NLP pipelines while incorporating advanced ontology-based reasoning, thereby bridging the gap between basic entity recognition and fully integrated ontology mapping.

(2) Reasoning Capabilities

Integrating reasoning capabilities into NLP pipelines is crucial for transforming raw text into actionable insights, particularly in complex analytical tasks where understanding context and drawing inferences are key. GraphRAG exemplifies this importance by not only extracting semantic relations but also providing detailed, interpretive explanations that reveal hidden connections and support the derivation of new knowledge. This advanced reasoning capability enhances decision-making and analytical precision in scenarios where mere linguistic parsing is insufficient.

In contrast, while spaCy and CoreNLP excel at processing language semantics and syntactic structures, they lack inherent reasoning mechanisms. To leverage these tools for inferential tasks, a multi-stage process can be implemented. One possible approach is as follows:

**Post-Processing Pipeline:**

1. *Extraction Phase:* Use spaCy/CoreNLP to perform named entity recognition, dependency parsing, and relation extraction.
2. *Graph Construction:* Map the extracted entities and relations onto a graph structure.
3. *Inference Module:* Integrate a reasoning engine—either a rule-based system or a neural logic module—that operates on the graph. (This module applies domain-specific rules or learned patterns to deduce implicit relationships and infer new knowledge from the structured data.)
4. *Explanation Generation:* Generate detailed explanations for each inferred relation, thereby enhancing the interpretability of the reasoning process. These explanations can be tied back to the original text, offering traceability and context for each inference.

### 4.1.5. Ease of Integration

(1) Compatibility with Other Tools

spaCy offers seamless integration, requiring minimal setup: users simply create an object, define functions to pass data, and can quickly generate triplets for graph-based analysis. This ease of integration makes spaCy highly compatible with existing systems, including the ability to be used as part of larger programs that call tools or functions, such as agents. Additionally, spaCy is an open-source library, which facilitates customization and integration with other open-source models. However, it does not inherently support third-party APIs, though users can easily wrap spaCy functions into REST APIs using frameworks like Flask or FastAPI.

In contrast, CoreNLP presents integration challenges due to the need for a properly functioning Java server, which must be accessed via API requests. This additional complexity can cause occasional integration issues, particularly when setting up the server or handling API interactions. While CoreNLP can be used in larger programs, it might require some additional effort to integrate. CoreNLP is also open-source.

GraphRAG strikes a balance, allowing for straightforward integration with other tools, necessitating only a few processing steps to extract the final answer from the returned text. GraphRAG can be used within agents to call tools or functions as part of a larger workflow, leveraging its ability to automate the extraction of a rich knowledge graph from text documents. While GraphRAG itself is not an open-source model in the traditional sense, it uses large language models that can be accessed via APIs, making it compatible with both open-source and proprietary models. Its integration with third-party APIs is facilitated by its deployment on platforms like Azure, which simplifies the interaction with other tools and service.

(2) API Availability

spaCy supports local usage on a CPU, with the option to run a containerized HTTP API available on Docker Hub, allowing users to deploy the tool as an API with minimal setup. CoreNLP, however, requires the setup of a dedicated Java server, which must be accessed via POST requests, adding complexity to the process of using it as an API. In contrast, GraphRAG simplifies the API setup, enabling users to run it as an API by executing a few command-line interface (CLI) commands, facilitating smoother integration with other systems.

Table 7: Comparison table for the 'API Availability' subparameter

| Feature | spaCy | CoreNLP | GraphRAG |
|---|---|---|---|
| **API Deployment** | Available via Docker as a containerized HTTP API | Requires manual setup of a Java-based server | Can be deployed via CLI with minimal setup |
| **Setup Complexity** | Simple; installing the package and running a few commands | Complex; setting up a server, handling configurations, and sending POST requests | Very simple; can be launched with a few CLI commands |
| **Dependency Requirements** | Python-based | Requires Java and additional configuration | Python-based |
| **Server Requirement** | Not required for local usage; API runs on demand | Requires a persistent Java-based server | No dedicated server needed; API runs dynamically |
| **Ease of Integration** | High; supports direct Python API calls and containerized deployment | Moderate; requires Java server setup and external API calls | High; easily integrates with Python-based workflows |
| **Startup Time for API** | Fast; initializes within seconds | Slow; server initialization takes additional time | Fast; minimal startup time due to lightweight architecture |

### 4.1.6. Hardware Requirements

(1) CPU and GPU Utilization, which assess the efficiency of resource usage for processing queries,

spaCy is optimized for CPU and GPU usage, offering small, medium, and large models tailored for efficient CPU processing. Written in Cython and utilizing its own Thinc library, spaCy also supports models from other libraries via wrappers. GPU utilization is supported through the CuPy module, which is compatible with various CUDA-enabled GPUs, with the transformer (trf) pipeline being particularly suited for GPU use. The tool can be switched to require GPU computation for enhanced performance, with the trf version consuming approximately 2 GB of RAM when run on Colab. CoreNLP, while not explicitly optimized for GPU usage, primarily targets CPU processing. Although it can run efficiently with enough space, its CPU-focused design limits GPU performance. GraphRAG, on the other hand, demands significant processing power when running sentence completion or embedding models locally. Due to multiple prompts processed at different stages, the sequential nature of its API calls makes it a resource-intensive tool, particularly in terms of computational cost associated with the large number of API calls during execution.

Table 8: Comparison table for the 'CPU/GPU Utilization' subparameter

| Feature | spaCy | CoreNLP | GraphRAG |
|---|---|---|---|
| **CPU Optimization** | Highly optimized for CPU; supports small, medium, and large models | Optimized for CPU usage; performs well on systems with ample CPU resources | Can run on a CPU but is less efficient |
| **GPU Support** | Supported via CuPy (for CUDA-enabled GPUs); the transformer pipeline benefits the most | No explicit GPU optimization; GPU usage is limited | Heavy GPU usage if locally run models are used |
| **RAM Usage** | ~2 GB RAM (transformer pipeline on Colab) | Varies based on pipeline; no clear GPU benefit | High – multiple prompts and API calls increase load |
| **Overall Hardware Demand** | Moderate – efficient on CPU, better with GPU for transformers | Low to Moderate – CPU-focused, limited GPU leverage | High – resource-intensive due to several API calls and embeddings |

(2) Cloud Compatibility

spaCy is inherently compatible with cloud environments due to its straightforward pipeline and model setup, making it easy to deploy on the cloud. Additionally, the availability of its API further simplifies its integration into cloud-based systems. CoreNLP requires more setup for cloud deployment, as it first establishes its own Java server, which then needs to be accessed via POST requests. Despite this added complexity, it can be configured to run as a cloud API after some integrations. GraphRAG is built for cloud use, with the preferred mode being to run its models or APIs in the cloud due to the significant GPU resources required for its models.

Table 9: Comparison table for the 'Cloud Compatibility' subparameter

| Feature | spaCy | CoreNLP | GraphRAG |
|---|---|---|---|
| **Ease of Cloud Deployment** | High – Minimal setup required due to pre-configured pipelines and models. | Moderate – Requires setting up a Java server before deployment. | High – Designed for cloud-first use with API-based interaction. |
| **Scalability** | Highly scalable; optimized for low-latency NLP tasks. | Can scale but requires proper backend management. | Scales well but demands high computational resources. |
| **Deployment Complexity** | Simple – Pre-trained models allow quick cloud integration. | Moderate – Additional server setup required before use. | Complex – High computational needs necessitate cloud infrastructure. |

### 4.1.7. Explainability & Transparency

spaCy offers high transparency, as the model decisions are based on a fixed flow within well-defined pipelines. This structured design ensures that every processing step—from tokenization to dependency parsing—is explicitly defined, making it easy to trace how graph triplets are extracted and how answers are generated. Since spaCy's performance is directly tied to the quality of training data and rule-based heuristics, its outputs are predictable and interpretable.

CoreNLP follows a similar approach, maintaining a transparent pipeline where each module, such as NER and OpenIE, contributes to graph formation in a well-defined manner. The explicit rules and linguistic patterns used in CoreNLP allow users to analyze which nodes and relations influence the final answer. This interpretability ensures that modifications to preprocessing steps or extraction rules lead to predictable changes in the generated knowledge graph and, subsequently, the LLM-generated responses.

In contrast, GraphRAG builds a knowledge graph by first extracting entities and relationships (often represented as triplets) from text using both rule-based methods and LLM-powered extraction. These triplets are then structured into nodes and edges, where key entities or relations (for example, those with high centrality or strong semantic relevance) may be prioritized through weighting schemes like similarity scores or inverse TF-IDF. Although GraphRAG's pipeline—from graph formation to prompt construction—is well-defined, the actual generation of answers is driven by an underlying probabilistic LLM. This introduces an extra layer of complexity that makes it more challenging to fully trace and explain how each output was derived, as the rationale behind the LLM's decisions is less interpretable compared to rule-based pipelines.

### 4.1.8. Licensing and Cost

spaCy is open-sourced under the MIT license, with no cost for the software except for the hardware required to run it. Additionally, spaCy offers commercial licenses and support through Explosion AI, which also provides Prodigy, a tool for fast annotation and iteration, along with the Thinc machine learning library. CoreNLP is open-sourced under the GNU-GPL license, allowing free use for non-commercial purposes, though it restricts use in proprietary software. For those requiring commercial usage, CoreNLP offers a commercial license for distributors utilizing proprietary software. GraphRAG is also open-sourced under the MIT license, providing full access to the framework and the freedom to use it without restrictions. The inherent cost associated with GraphRAG depends on the open-source nature of the underlying LLM model, which may vary based on the specific model used. It can be utilized for free if the LLM model is open-source, providing an open-source benefit. However, the cost associated with LLMs is to be considered for the other two methods as well. So, that part of the cost is common for all three methods.

Table 10: Comparison table for the 'Licensing and Cost' subparameter

| **Feature** | **spaCy** | **CoreNLP** | **GraphRAG** |
|---|---|---|---|
| **License Type** | MIT License (Permissive) | GNU-GPL License (Restrictive) | MIT License (Permissive) |
| **Commercial Restrictions** | No restrictions; can be used in proprietary software | Cannot be used in proprietary software without a commercial license | No restrictions whatsoever |
| **Cost** | Free for all uses; optional paid support available | Free for non-commercial use; paid license required for commercial applications | Totally free for all uses |
| **Ease of Integration in Proprietary Software** | Easy; no restrictions on modification or integration | Restricted; requires a commercial license for proprietary software | Easy; no restrictions on use, merge, modification, or distribution |

## 4.2. The Performance-Based Comparative Study

### 4.2.1. Evaluations

4.2.1.1. "All's Well That Ends Well"

The following were the four questions that were asked.
Q1: Describe the relationship dynamics between Helena and Bertram.
Q2: Can you analyze the evolution of Helena and Bertram's relationship?
Q3: Examine the complexities of Helena and Bertram's relationship.
Q4: Explore the nuances of Helena and Bertram's relationship.

All answers are graded by an expert human evaluator and a GPT-4 model based on a given rubric without letting them know which of the methods has produced any particular answer. The answers are also generated in two ways: one by informing the model that the answer will be evaluated based on the provided rubric, and one without informing it.

Refer to the Appendix for the detailed prompts used by the answering LLM and the evaluator LLM.

The following tables consist of the overall scores the model and evaluator gave in the following two scenarios. The expert and the GPT-4 evaluator both gave a score from A, B, C, and D on each of the parameters. Then, firstly, the grades are mapped with numbers A–5, B–4, C–3, and D–2. After that, the overall score for each of the responses is calculated by averaging the four individual fronts' scores, which is done to ensure that the responses, hence approaches, are evaluated more subjectively.

**Scenario 1: Rubric not given to the responding LLM**

*Scores by Expert evaluator*

Table 11: Expert Evaluator scores (without rubric)

| Method | Q1 | Q2 | Q3 | Q4 |
|---|---|---|---|---|
| spaCy | 4 | 4.25 | 4.25 | 4.25 |
| CoreNLP | 4.5 | 4 | 4.25 | 3.75 |
| GraphRAG | 4.25 | 4.5 | 4.75 | 4.5 |

*Scores by GPT-4 evaluator*

Table 12: GPT-4 Evaluator scores (without rubric)

| Method | Q1 | Q2 | Q3 | Q4 |
|---|---|---|---|---|
| spaCy | 4.5 | 3.5 | 5 | 4.75 |
| CoreNLP | 5 | 4.75 | 4.75 | 5 |
| GraphRAG | 4.25 | 5 | 5 | 5 |

**Scenario 2: Rubric given to the responding LLM**

*Scores by Expert evaluator*

Table 13: Expert Evaluator scores (with rubric)

| Method | Q1 | Q2 | Q3 | Q4 |
|---|---|---|---|---|
| spaCy | 4.25 | 4.5 | 4.25 | 4.25 |
| CoreNLP | 3.5 | 3.75 | 4 | 4.25 |
| GraphRAG | 4.75 | 4.25 | 4.5 | 4.25 |

*Scores by GPT-4 evaluator*

Table 14: GPT-4 Evaluator scores (with rubric)

| Method | Q1 | Q2 | Q3 | Q4 |
|---|---|---|---|---|
| spaCy | 5 | 5 | 5 | 5 |
| CoreNLP | 3.5 | 5 | 5 | 5 |
| GraphRAG | 5 | 5 | 5 | 5 |

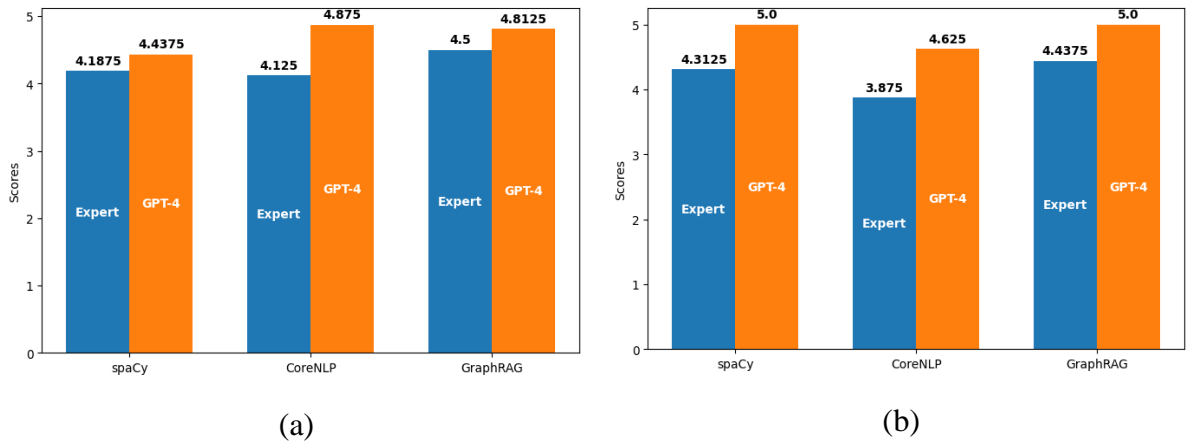

Figure 4: Scores comparison (a) Without rubric scores, (b) With rubric scores

\* The detailed responses provided by each method are given in the appendix.

The experimental results reveal that among the three approaches tested, GraphRAG consistently outperforms the others. On average, GraphRAG achieved scores of 4.46 from the expert evaluator and 4.90 from GPT-4, clearly indicating its superiority in generating effective responses. In contrast, the spaCy+LLM approach received scores of 4.25 (expert) and 4.71 (GPT-4), while the CoreNLP+LLM approach obtained slightly lower scores of 4.00 (expert) and 4.75 (GPT-4). Although both evaluators concurred that GraphRAG produced the most effective outputs, their relative assessments of the spaCy and CoreNLP-based methods differed.

A deeper look into the comparative performance of spaCy+LLM and CoreNLP+LLM reveals interesting nuances. The expert evaluator rated spaCy+LLM slightly higher than CoreNLP+LLM (4.25 vs. 4.00), whereas GPT-4 assigned nearly identical scores to both approaches. This divergence suggests that while both systems can produce valuable outputs, there are underlying differences in how each approach structures and presents information. The human expert might be more sensitive to the subtleties of information organization and clarity, favoring the outputs from spaCy+LLM over those from CoreNLP+LLM.

The discrepancy in scores becomes even more intriguing when considering that CoreNLP produced a significantly larger number of triplets from each text compared to spaCy. While assuming that more data leads to better performance might be intuitive, our findings suggest otherwise. An overabundance of extracted information can overwhelm the language models, leading to diminished performance in generating a coherent, focused response. In this case, the surplus of triplets may introduce redundancy or even noise, which could impair the LLM's ability to prioritize the most salient details, ultimately resulting in a lower or comparable score despite the larger volume of data.

Another noteworthy inference is the consistent trend of GPT-4, which has awarded higher scores than the human expert. This observation implies that GPT-4 might have an inherent bias toward more generous evaluations, potentially due to its training on data that emphasizes positive reinforcement or broader acceptance of varied response forms. While this bias does not invalidate the assessments, it highlights the importance of calibrating automated evaluations to align more closely with human judgment, especially when these systems are used as proxies for expert evaluation.

Lastly, the statistically significant correlation between the expert and GPT-4 scores—reflected by a Pearson correlation coefficient of 0.415 and a p-value of 0.043—underscores a meaningful relationship between human and automated assessments. This correlation indicates that, despite some differences in absolute scoring levels, GPT-4 is trying to capture similar trends in response quality as human experts. This indicates that even though the GPT-4 evaluator tries well to grade the answers, it is still a bit too far away from replacing the expert evaluator altogether. As this gap between automated and human evaluation narrows, there is promising potential for LLMs to assist or even replace human evaluators in certain contexts, thereby reducing workload and improving efficiency in large-scale evaluation tasks.

4.2.1.2. "RepliQA"

Ten documents selected at random, with the character length being between 3000 to 4000. Each of the documents has 5 different questions associated, which have to be strictly answered using the document context provided. Answers are then generated using all three methods. Again, for each method, the answers are generated once by letting the model know about the rubric to be used for evaluation and once without being informed about the rubric.

The following table indicates the scores obtained by the methods, using different metrics, which are often used for scoring the performance in QA tasks.

Table 15: Metric-based Evaluation of different methods (RepliQA dataset)

| Metrics | Spacy | | CoreNLP | | GraphRAG | |
|---|---|---|---|---|---|---|
| | W/o rubric | With rubric | W/o rubric | With rubric | W/o rubric | With rubric |
| Exact Match | 0 | 0 | 0 | 0 | 0 | 0 |
| QA F1 (squad) | 17.06708 | 10.27871 | 16.00828 | 10.10424 | 22.62497 | 15.60035 |
| ROUGE-L F1 | 0.12015 | 0.07654 | 0.11541 | 0.07786 | 0.17989 | 0.12274 |
| METEOR | 0.17908 | 0.18583 | 0.17600 | 0.18639 | 0.32620 | 0.26990 |
| BERTScore Precision | 0.83116 | 0.79125 | 0.83045 | 0.79270 | 0.80961 | 0.79265 |
| BERTScore Recall | 0.85155 | 0.84212 | 0.85037 | 0.84119 | 0.87217 | 0.86168 |
| BERTScore F1 | 0.84066 | 0.81569 | 0.83966 | 0.81604 | 0.83949 | 0.82552 |

**Exact Match (EM):** None of the three methods (spaCy, CoreNLP, GraphRAG) achieved any Exact Match (EM) scores, even with rubric augmentation. This indicates generated answers didn't precisely match ground truth responses, likely due to lexical variability and EM's strictness. More semantic metrics like F1 or BERTScore are needed for evaluation.

**QA F1 (SQuAD):** GraphRAG significantly outperformed spaCy and CoreNLP in QA F1 scores (22.62 without rubric, 15.60 with rubric). Interestingly, rubric integration decreased scores for all methods, possibly due to added context complicating concise answer retrieval. GraphRAG's superiority may stem from better knowledge structure integration, yielding contextually aligned, non-rubric-dependent answers.

**ROUGE-L F1:** GraphRAG outperforms spaCy and CoreNLP in ROUGE-L (0.179 w/o rubric, 0.122 with), despite rubrics lowering scores across all systems. Similar to QA F1, adding rubric context resulted in a drop across all systems, albeit less sharply. This implies that while rubrics provide guidance, they might steer the answer generation away from the ground-truth phrasing, leading to reduced lexical overlap. GraphRAG's edge suggests superior contextual encoding and retrieval, even when rubrics don't align with expected phrasing.

**METEOR:** METEOR scores for spaCy and CoreNLP remained stable or slightly improved with rubric addition. GraphRAG scored the highest, however, saw a dip from 0.326 to 0.269 on addition of rubric, suggesting its rubric-conditioned outputs diverge semantically from ground truth, unlike spaCy and CoreNLP's smaller fluctuations. This suggests that GraphRAG's more flexible output, when constrained by rubric, can overshoot the semantic target.

**BERTScore Precision:** BERTScore Precision for spaCy and CoreNLP (0.83) is higher than for GraphRAG (0.81). Rubric addition slightly decreased precision for all, suggesting that while rubrics add context, they also introduce noise, impacting directness.

**BERTScore Recall:** GraphRAG outperforms spaCy and CoreNLP in BERTScore Recall (with/without rubric, ~0.87/~0.86), showing better information capture from ground truth. High recall across systems suggests answers cover key content. Slight recall decline with rubric implies verbosity or less focused content, despite retaining answer components.

**BERTScore F1:** All three methods show comparable BERTScore F1 (~0.84 without rubric), with GraphRAG slightly more robust with rubric. While rubric-augmented answers may be less precise, their smaller BERTScore F1 reduction suggests they maintain semantic alignment, potentially aiding semantic generalization. This also highlights GraphRAG's ability to generate semantically coherent, context-rich answers.

Then, for the next part of the experimentation, just as it was done for Shakespeare's play, the responses were evaluated by the GPT-4 model, which had rubric with itself to grade them. The following table summarizes the scoring that GPT-4 gave for different methods and scenarios as below.

Table 16: GPT-4 Evaluator scores (RepliQA dataset)

| Method/Scenario | Average Score |
|---|---|
| Given Dataset Answer | 3.71 |
| spaCy (without rubric) | 4.18 |
| spaCy (with rubric) | 4.655 |
| CoreNLP (without rubric) | 4.165 |
| CoreNLP (with rubric) | 4.655 |
| GraphRAG (without rubric) | 4.61 |
| GraphRAG (with rubric) | 4.66 |

According to the GPT-4 evaluator scores on the RepliQA dataset, all the generated responses outperformed the dataset's original answers, as the models generated bigger and fancier answers which scored higher when compared against straight up to the point ground truth answers, due to the way, the rubric is defined to score the answers. Among the approaches, GraphRAG (with rubric) achieved the highest average score of 4.66, closely followed by CoreNLP (with rubric) and spaCy (with rubric), all at 4.655, suggesting that these systems generated responses of nearly equivalent quality when evaluated holistically by a large language model. Notably, CoreNLP and spaCy benefited more from rubric augmentation, implying that rubric guidance helped compensate for their relatively limited contextual reasoning. In contrast, GraphRAG's score remained almost unchanged regardless of rubric usage, indicating minimal sensitivity to additional rubric context.

GraphRAG demonstrated consistently high performance with and without rubric, reflecting its strength in integrating structured knowledge and contextual cues. The small yet positive gain from rubric inclusion (4.61 to 4.66) further supports its capacity to incorporate guidance effectively. Overall, these results not only underscore GraphRAG's superiority but also suggest that rubric augmentation can enhance model outputs, particularly for models that can flexibly integrate external guidance.

Then, all of the responses for a particular question, along with the question and the context were given to the GPT-4 model, and asked to relatively rank the responses from best to worst. The following table shows how many times the respective responses were ranked first or second in the rankings given by GPT-4.

| Method | Rank 1 Count | Rank 2 Count |
|---|---|---|
| **Dataset Answer** | 9 | 6 |
| **SpaCy (without rubric)** | 15 | 3 |
| **SpaCy (with rubric)** | 9 | 12 |
| **CoreNLP (without rubric)** | 0 | 4 |
| **CoreNLP (with rubric)** | 2 | 9 |
| **GraphRAG (without rubric)** | 4 | 8 |
| **GraphRAG (with rubric)** | 11 | 8 |

The rank-based comparison highlights an interesting distinction between spaCy and GraphRAG. SpaCy (without rubric) secures the highest number of rank-1 placements (15), suggesting that it occasionally produces responses that are judged as the best in isolated comparisons. However, when evaluated in terms of overall reliability, GraphRAG demonstrates greater consistency. GraphRAG with rubric, for example, secures 11 rank-1s and 8 rank-2s (19 in total), reflecting a balance of strong answers without significant drops in quality. When considered alongside the average score results, it becomes clear that while spaCy excels in producing standout answers on occasion, GraphRAG achieves a more stable and dependable performance across the dataset.

The effect of rubric use is also evident. For both spaCy and CoreNLP, introducing rubric guidance shifts performance: spaCy's rank-1 count drops from 15 to 9, but its rank-2 placements rise significantly from 3 to 12, indicating that rubric enforcement pushes its answers toward consistently "good" quality, even if fewer are rated as the absolute best. CoreNLP shows a similar trend, with rubric use improving its top-2 presence (from 4 to 11), though it still lags behind the other methods. GraphRAG, in contrast, benefits less dramatically from rubric introduction because it is already strong without rubric (12 top-2 placements versus 19 with rubric). This suggests that rubric guidance stabilizes weaker models (spaCy, CoreNLP) by reducing variability, whereas GraphRAG already exhibits high robustness and only gains marginally with rubric.

In summary, rubric-based evaluation enhances consistency across models, ensuring fewer poor-quality answers, but its effect is most pronounced for methods prone to variability. GraphRAG's relative insensitivity to rubric presence underscores its inherent robustness, while spaCy's shift from rank-1 dominance to rank-2 consistency reveals a trade-off between peak and average performance.

**NOTE:** It is important to note that comparing spaCy, CoreNLP, and GraphRAG on speed or memory usage is not meaningful due to their inherently different processing approaches—spaCy uses a lightweight Cython-based pipeline, CoreNLP employs a multi-stage Java framework with deep linguistic analysis, and GraphRAG leverages LLMs for hierarchical graph-based reasoning—making direct benchmarking of resource usage inherently unequal.

### 4.2.3. Pre-processing Requirements

The spaCy-based approach requires extensive manual preprocessing, including text segmentation, custom dependency parsing, and triplet extraction, demanding significant coding effort. The CoreNLP method automates many linguistic analyses through its OpenIE module but still necessitates server setup and some manual processing for question analysis and triplet matching. In contrast, GraphRAG automates the entire preprocessing pipeline, handling text ingestion, knowledge graph construction, and retrieval processes internally, thereby minimizing manual intervention. The following table summarizes the preprocessing requirements for each method:

Table 17: Comparison table for the 'Pre-processing Requirements' subparameter

| **Aspect** | **spaCy** | **CoreNLP** | **GraphRAG** |
|---|---|---|---|
| **Setup complexity** | Low | Medium (Java server required) | Very low |
| **Dependency parsing** | Custom patterns & rules | Built-in | Fully automated |
| **Triplet extraction** | Hand-rolled via patterns | OpenIE module | Internal to GraphRAG (using LLMs) |
| **Question matching/filter** | Custom code | Custom code | Handled by framework (using prompts) |
| **Graph construction** | Manual | Manual | Automated |
| **Total manual effort** | High | Moderate | Low |

## 4.3. Added GraphRAG advantages

GraphRAG offers several advantages in the context of LLM + graph-based question answering. It excels at handling both structured and unstructured data, leveraging knowledge graphs to capture relationships between entities while extracting insights from unstructured text. This dual capability enhances response accuracy and comprehensiveness. Additionally, GraphRAG supports incremental learning, allowing for dynamic updates to the knowledge graph without requiring full retraining, which is particularly beneficial for rapidly evolving domains.

The system's schema flexibility allows for easy modification or extension of the underlying schema, supporting dynamic or schema-less environments. Furthermore, GraphRAG is designed for distributed processing, leveraging cloud-native architectures to ensure scalability and high performance. GraphRAG also provides provenance tracking, tracing the origin of each piece of information used in generating responses. This feature enhances transparency and trustworthiness by grounding outputs in structured formats, reducing hallucinations and providing confidence scoring. Overall, GraphRAG's comprehensive capabilities make it a powerful tool for complex AI applications across various domains, offers a robust solution for question answering, ensuring privacy, and supporting advanced reasoning and processing capabilities.

## 5. Conclusion and Future Work

In this work, we have presented a fully reproducible framework that leverages three distinct open-source methods—spaCy, Stanford CoreNLP-OpenIE, and GraphRAG—for extracting relational triplets, constructing knowledge graphs, and integrating them with large language models (LLMs) to enhance question-answering. Through systematic evaluation under identical conditions, we found that *spaCy* delivers a lightweight, high-precision baseline: it excels at extracting clear, well-formed subject–predicate–object triplets with minimal false positives, yet its coverage diminishes when faced with idiomatic or complex sentence structures. *CoreNLP-OpenIE*, by contrast, achieves the broadest factual coverage across varied text types, uncovering nuanced relations even in less canonical phrasing; however, this comes at the expense of higher computational overhead and the need for more aggressive filtering to remove noisy or overly granular triplets. *GraphRAG* stands out for its reasoning prowess: its tight coupling of graph-structured context with LLM prompting yields the richest, most coherent responses on thematic and multi-hop queries, albeit with added system complexity and increased runtime demands. Our empirical results—validated by GPT-4 automated scoring and domain expert judgments—demonstrate that each method occupies a unique point in the precision-coverage-reasoning trade-off space, thereby offering clear guidance on tool selection for diverse QA scenarios.

Based on these insights, we recommend a hybrid pipeline that begins with spaCy's high-precision filters to eliminate spurious relations, followed by CoreNLP-OpenIE's expansive extraction to maximize recall and culminates with GraphRAG's deep contextual reasoning to generate the final answer. To support real-time and dynamic applications, knowledge graphs should be incrementally updated via streaming triplet ingestion, ensuring that the QA system remains aligned with evolving information. Incorporating lightweight expert-in-the-loop validations—such as crowd-sourced or calibrated domain checks—can further prune residual noise and calibrate LLM prompts for heightened coherence. Finally, practitioners should tune extraction thresholds and prompt templates to the specific demands of their domain, balancing latency, accuracy, and interpretability according to task requirements.

Looking ahead, a promising research avenue lies in unifying triplet extraction and reasoning within end-to-end hybrid architectures. One possible avenue can be where the LLMs are designed such that building knowledge graphs is a part of their own learning process, using trainable components that can be adjusted during training. This way, the model can automatically learn which relationships and structures are most helpful for answering questions. This could eliminate the need for separate, multi-step RAG pipelines, making the system more efficient and integrated. A key question for future work is whether graph-centric retrieval—directly querying over learned graph embeddings—can fully supplant traditional chunk-based RAG methods without sacrificing multi-hop inference capabilities or explainability. To date, our study has been anchored in text-centric extraction; expanding this comparison to include retrieval over vectorized graph representations will yield valuable insights into the fundamental limits of RAG versus graph-based strategies.

Beyond retrieval and model architecture, scalable evaluation and multimodal integration represent critical frontiers. Automated, graph-based coherence metrics—such as entailment scoring over relation paths—could reduce our reliance on costly expert assessments, while calibrated crowd-sourcing frameworks can facilitate large-scale benchmarking. On the content side, extending knowledge graphs to encapsulate visual, tabular, and temporal data will enable QA systems to operate over charts, images, and time series, unlocking new applications in scientific analysis, financial forecasting, and beyond. Finally, developing domain-specific schema adaptation and incremental graph-updating strategies will ensure that KG-enhanced QA systems remain robust and current across rapidly evolving knowledge domains. These combined efforts will bring us closer to QA systems capable of human-level reasoning, contextual richness, and real-world applicability.

## 7. Appendix

**Prompt for the answering LLM:-**

> You are an expert writer tasked with crafting an answer that will be evaluated based on the following four factors: **Content/Ideas**, **Organization**, **Style**, and **Mechanics**. Your goal is to create a response that would achieve the highest score possible (A) in each of these categories.
>
> **Writing Criteria for Top Scores (A):**
>
> 1. **Content/Ideas**:
>    - Demonstrate deep understanding and thorough analysis of the question.
>    - Present insightful, original, and well-developed ideas with comprehensive detail.
>    - Address all parts of the question fully, providing evidence or examples as needed.
>    - Ensure the response follows all literary or academic conventions relevant to the topic.

2. **Organization**:
    - Structure the response logically, with a clear introduction, well-organized body paragraphs, and a strong conclusion.
    - Ensure smooth transitions between ideas, keeping the answer focused and coherent.
    - Integrate any quotations, examples, or evidence seamlessly into the discussion.
3. **Style**:
    - Write in an engaging, sophisticated tone, using varied sentence structures and precise, appropriate vocabulary.
    - Enhance the clarity and impact of the content through effective word choice and sentence flow.
    - Follow the appropriate citation style (e.g., MLA format) without errors.
4. **Mechanics**:
    - Ensure your response is free from grammar, punctuation, and spelling errors.
    - Polish the writing so it is clean and professional, with no distracting mistakes.

**Instructions for Writing the Response:**

1. Carefully read the question and think about how to address each part fully, while adhering to the highest standards for each of the four factors mentioned.
2. Organize your thoughts clearly before you begin writing, ensuring a logical flow of ideas.
3. Use a refined and varied writing style to make your response engaging and clear.
4. Proofread your answer to eliminate any errors in grammar, punctuation, or spelling.

Write your response to the following question, keeping in mind the criteria for achieving an "A" in **Content/Ideas**, **Organization**, **Style**, and **Mechanics**:

**Question:**

**Evaluator Prompt:-**

You are an expert evaluator responsible for grading responses based on four factors: **Content/Ideas**, **Organization**, **Style**, and **Mechanics**. Your task is to assign a grade (A, B, C, or D) for each factor and provide an overall rating with a brief explanation.

**Evaluation Criteria:**

1. **Content/Ideas**:
    - **A**: Demonstrates deep understanding, insightful ideas, and thorough analysis. Fresh, original concepts with comprehensive coverage of the question.
    - **B**: Shows solid understanding, with good ideas but may lack depth or originality. Mostly accurate but less detailed.
    - **C**: Shows partial understanding. Ideas are incomplete, superficial, or somewhat inaccurate. Some gaps in logic or evidence.
    - **D**: Lacks meaningful ideas or relevant content. Large parts of the question are incorrect or missing.

2. **Organization**:
   - **A**: Logically organized with smooth transitions and a strong introduction and conclusion. All ideas are relevant and well-integrated.
   - **B**: Generally organized but may lack clarity in some transitions or sections. The structure is clear, but the introduction or conclusion could be stronger.
   - **C**: Basic structure with unclear flow. Ideas may jump around or lack proper transitions.
   - **D**: Disorganized and difficult to follow, with little to no logical flow or coherence.
3. **Style**:
   - **A**: Engaging and sophisticated tone, varied sentence structure, and appropriate vocabulary. The style enhances clarity.
   - **B**: Consistent tone and vocabulary, though lacks stylistic variety. Clear but not as engaging.
   - **C**: Basic and repetitive, lacking refinement. Understandable but not compelling.
   - **D**: Awkward or unclear, with inappropriate tone or misused vocabulary.
4. **Mechanics**:
   - **A**: Minimal or no grammar, punctuation, or spelling errors.
   - **B**: A few minor errors, but not distracting.
   - **C**: Some significant errors, occasionally distracting.
   - **D**: Frequent and severe errors, making it hard to understand.

**Grading Instructions:**

1. Review the response using the four factors above.
2. Assign a grade (A, B, C, or D) for each factor.
3. Provide an overall rating with a brief explanation based on the overall performance across all factors.

**Expected Output:**

- **Content/Ideas**: B
- **Organization**: A
- **Style**: B
- **Mechanics**: A
- **Overall Rating**: B
  **Explanation**: The response shows a good understanding with clear organization. However, some ideas are underdeveloped, preventing an A in Content/Ideas, and minor stylistic improvements could make the response more engaging.

**Detailed grading for each of the approaches for the Shakespeare's play:-**

Scenario 1: Rubric not given to the responding LLM

*Grades by Expert evaluator for spaCy-based approach*

| Evaluation criteria | Q1 | Q2 | Q3 | Q4 |
| --- | --- | --- | --- | --- |
| **Content/Ideas** | B | A | A | A |
| **Organization** | B | B | B | B |
| **Style** | B | B | B | B |
| **Mechanics** | B | B | B | B |
| **Overall** | 4 | 4.25 | 4.25 | 4.25 |

*Grades by GPT-4 evaluator for spaCy-based approach*

| Evaluation criteria | Q1 | Q2 | Q3 | Q4 |
|---|---|---|---|---|
| **Content/Ideas** | B | C | A | A |
| **Organization** | A | B | A | A |
| **Style** | B | C | A | B |
| **Mechanics** | A | B | A | A |
| **Overall** | 4.5 | 3.5 | 5 | 4.75 |

*Grades by Expert evaluator for CoreNLP-based approach*

| Evaluation criteria | Q1 | Q2 | Q3 | Q4 |
|---|---|---|---|---|
| **Content/Ideas** | A | B | A | B |
| **Organization** | B | B | B | B |
| **Style** | B | B | B | C |
| **Mechanics** | A | B | B | B |
| **Overall** | 4.5 | 4 | 4.25 | 3.75 |

*Grades by GPT-4 evaluator for CoreNLP-based approach*

| Evaluation criteria | Q1 | Q2 | Q3 | Q4 |
|---|---|---|---|---|
| **Content/Ideas** | A | A | B | A |
| **Organization** | A | A | A | A |
| **Style** | A | B | A | A |
| **Mechanics** | A | A | A | A |
| **Overall** | 5 | 4.75 | 4.75 | 5 |

*Grades by Expert evaluator for GraphRAG-based approach*

| Evaluation criteria | Q1 | Q2 | Q3 | Q4 |
|---|---|---|---|---|
| **Content/Ideas** | A | A | A | A |
| **Organization** | B | B | A | A |
| **Style** | B | B | B | B |
| Mechanics | B | A | A | B |
| Overall | 4.25 | 4.5 | 4.75 | 4.5 |

*Grades by GPT-4 evaluator for GraphRAG-based approach*

| Evaluation criteria | Q1 | Q2 | Q3 | Q4 |
|---|---|---|---|---|
| Content/Ideas | B | A | A | A |
| Organization | A | A | A | A |
| Style | B | A | A | A |
| Mechanics | B | A | A | A |
| Overall | 4.25 | 5 | 5 | 5 |

Scenario 2: Rubric given to the responding LLM

*Grades by Expert evaluator for spaCy-based approach*

| Evaluation criteria | Q1 | Q2 | Q3 | Q4 |
|---|---|---|---|---|
| Content/Ideas | A | A | A | A |
| Organization | B | B | B | B |
| Style | B | A | B | B |
| Mechanics | B | B | B | B |
| Overall | 4.25 | 4.5 | 4.25 | 4.25 |

*Grades by GPT-4 evaluator for spaCy-based approach*

| Evaluation criteria | Q1 | Q2 | Q3 | Q4 |
|---|---|---|---|---|
| Content/Ideas | A | A | A | A |
| Organization | A | A | A | A |
| Style | A | A | A | A |
| Mechanics | A | A | A | A |
| Overall | 5 | 5 | 5 | 5 |

*Grades by Expert evaluator for CoreNLP-based approach*

| Evaluation criteria | Q1 | Q2 | Q3 | Q4 |
|---|---|---|---|---|
| Content/Ideas | C | A | A | A |
| Organization | C | C | C | B |
| Style | B | C | B | B |
| Mechanics | B | B | B | B |

| | Overall | 3.5 | 3.75 | 4 | 4.25 |
|---|---|---|---|---|---|

*Grades by GPT-4 evaluator for CoreNLP-based approach*

| Evaluation criteria | Q1 | Q2 | Q3 | Q4 |
|---|---|---|---|---|
| Content/Ideas | D | A | A | A |
| Organization | C | A | A | A |
| Style | A | A | A | A |
| Mechanics | B | A | A | A |
| Overall | 3.5 | 5 | 5 | 5 |

*Grades by Expert evaluator for GraphRAG-based approach*

| Evaluation criteria | Q1 | Q2 | Q3 | Q4 |
|---|---|---|---|---|
| Content/Ideas | A | B | A | B |
| Organization | B | B | B | B |
| Style | A | A | B | B |
| Mechanics | A | B | A | A |
| Overall | 4.75 | 4.25 | 4.5 | 4.25 |

*Grades by GPT-4 evaluator for GraphRAG-based approach*

| Evaluation criteria | Q1 | Q2 | Q3 | Q4 |
|---|---|---|---|---|
| Content/Ideas | A | A | A | A |
| Organization | A | A | A | A |
| Style | A | A | A | A |
| Mechanics | A | A | A | A |
| Overall | 5 | 5 | 5 | 5 |

**Responses generated by each of the approaches for the Shakespeare's play:-**

spaCy-based approach

| Questions | Without Rubric response | With rubric response |
|---|---|---|
| Q1. | In William Shakespeare's play "All's Well That Ends Well," Helena and Bertram have a complex and tumultuous relationship dynamic. Bertram, a nobleman, is initially unwilling to marry Helena, a low-born woman, due to her social status and his own pride. However, Helena's love and determination ultimately win him over, and they get married.\n\nThroughout the play, Helena's love for Bertram is depicted as unwavering and pure, while Bertram's feelings towards Helena are more ambivalent. He is initially reluctant to reciprocate her love and is dismissive of her advances. Bertram's behavior is motivated by his own sense of pride and social hierarchy, as he views Helena as being beneath him.\n\nDespite Bertram's initial reluctance, Helena's devotion and perseverance eventually break down his defenses, and he begins to develop feelings for her. However, their relationship is still marked by tension and conflict, particularly due to Bertram's lingering sense of superiority and Helena's own insecurities.\n\nUltimately, the play suggests that true love can conquer even the most entrenched social barriers, and that relationships can be redeemed through mutual understanding, respect, and devotion. Despite their differences, Helena and Bertram come to realize their love for each other and ultimately find happiness together. | **The Enigmatic Bond: Unraveling the Complex Relationship Dynamics between Helena and Bertram**\n\nIn William Shakespeare's All's Well That Ends Well, the intricate web of relationships between characters serves as a microcosm for the complexities of human connection. Among the most fascinating and enigmatic relationships in the play is that between Helena, the protagonist, and Bertram, the reluctant hero. Their bond is marked by a peculiar blend of affection, duty, and obligation, which Shakespeare skillfully explores throughout the play.\n\nAt the outset, Helena's love for Bertram appears as an all-consuming passion, driven by a deep sense of devotion and admiration. Her unwavering dedication to him is exemplified in her willingness to undertake the arduous journey to France to win his hand in marriage, despite his initial reluctance and dismissal of her affections. This unreciprocated love is rooted in Helena's genuine admiration for Bertram's noble lineage and her desire to elevate her social status by marrying into his family. As the play progresses, however, it becomes clear that Helena's love for Bertram is more than a mere infatuation; it is an integral part of her identity, and she is willing to risk everything to win his acceptance and affection.\n\nBertram, on the other hand, is a character plagued by his own insecurities and sense of duty. His reluctance to marry Helena stems from a desire to avoid the responsibility of settling down and a fear of being tied to a woman who is socially beneath him. Throughout the play, Bertram's behavior is marked by a sense of obligation, as he feels duty-bound to marry Helena in order to fulfill his father's wishes and secure his family's reputation. This sense of duty is at odds with his own desires and feelings, leading to a complex web of emotions that ultimately contributes to the tumultuous nature of his relationship with Helena.\n\nThe dynamic between Helena and Bertram is further complicated by the presence of other characters, including the Countess of Rossillion, Bertram's mother, and the King of France. The Countess, in particular, plays a significant role in shaping the relationship between Helena and Bertram, as she serves as a mediator and confidante to both characters. Her wisdom and guidance help to facilitate a deeper understanding between the two, ultimately leading to a reconciliation and a resolution to their conflicts.\n\nIn conclusion, the relationship dynamics between Helena and Bertram in All's Well That Ends Well are a rich tapestry of complex emotions, obligations, and desires. Shakespeare masterfully explores |
| Q2. | Helena and Bertram's relationship undergoes significant development throughout Shakespeare's play "All's Well That Ends Well". Here's a breakdown of their relationship's evolution:\n\n1. Initial encounter: The play opens with Helena, a gentlewoman, falling in love with Bertram, a young nobleman. Although Bertram is uninterested and even dismissive of Helena, she remains devoted to him.\n2. Bertram's rejection: When Helena's father, the Count of Rossillion, dies, Bertram inherits his title and fortune. Bertram, however, is unwilling to marry Helena, considering her social status inferior to his own. He rejects her and leaves the country, leaving Helena heartbroken.\n3. Helena's determination: Despite Bertram's rejection, Helena decides to follow him to Paris, where he has married the Princess of France, Helen. Helena's determination to win Bertram's love is evident in her willingness to undertake a treacherous journey and risk social shame.\n4. Helena's disguise: Helena disguises herself as a man (Diana) and becomes a member of the Princess's household. She uses this disguise to gain Bertram's trust and eventually becomes his page.\n5. Bertram's gradual affection: Over time, Bertram begins to develop feelings for Helena, albeit initially unaware of her true identity. He mistakes her for his page and becomes fond of her, eventually acknowledging her worth and intelligence.\n6. Revelation and reconciliation: The truth about Helena's identity is eventually revealed, and Bertram is forced to confront his earlier rejection. He comes to realize the error of his ways and apologizes to Helena, admitting his love for her.\n7. Marriage and resolution: The play concludes with Bertram and Helena getting married, with the blessing of the King of France. Bertram's transformation from a self-absorbed and ungrateful nobleman to a loving and appreciative husband marks a significant character development.\n\nThroughout the play, Helena's love for Bertram remains constant, while Bertram's feelings for her evolve from rejection to affection and eventually love. The evolution of their relationship serves as a commentary on the nature of love, social class, and personal growth. | **The Evolution of Helena and Bertram's Relationship: A Transformation from Misunderstanding to Mutual Understanding**\n\nIn William Shakespeare's play, All's Well That Ends Well, the complex and tumultuous relationship between Helena and Bertram undergoes a significant transformation, evolving from one marked by misunderstanding and rivalry to a deep and abiding connection. This analysis will examine the key events, character traits, and themes that contribute to the development of their relationship, ultimately demonstrating how their bond grows stronger through a process of mutual understanding and respect.\n\nInitially, Helena and Bertram's relationship is marred by a deep-seated misunderstanding. Bertram, the son of the Countess of Rossillion, is disdainful of Helena, a lowly gentlewoman, due to her perceived inferior social status. This prejudice is further exacerbated by Bertram's reluctance to marry Helena, whom he views as a mere servant. Conversely, Helena's love for Bertram is pure and unwavering, driven by her admiration for his noble character and her desire to elevate her social standing. This disparity in their perspectives leads to a rift between the two, as Bertram's indifference and Helena's devotion create a sense of tension and conflict.\n\nHowever, as the play progresses, a series of events and character developments contribute to a significant shift in their relationship. Bertram's banishment from France, following his refusal to marry Helena, serves as a wake-up call, forcing him to reevaluate his priorities and appreciate the value of true love. Meanwhile, Helena's determination and perseverance in pursuing Bertram demonstrate her unwavering commitment to their relationship, ultimately winning him over. This transformation is exemplified in Act 5, Scene 3, where Bertram, now aware of Helena's dedication and the depth of her love, begins to reciprocate her emotions, acknowledging her as his equal and his true companion.\n\nThe evolution of their relationship is also characterized by a growing mutual understanding and respect. As Helena and Bertram come to appreciate each other's strengths and weaknesses, they develop a deeper empathy and connection. Helena's kindness, compassion, and intelligence, initially overlooked by Bertram, are now recognized and valued by him. Similarly, Bertram's noble character, initially perceived as aloof and unyielding, is revealed to be tempered by a sense of honor and loyalty. This newfound understanding and respect serve as the foundation upon which their relationship is rebuilt, fostering a sense of trust and intimacy.\n\nFurthermore, the themes of love, identity, and social class play a crucial role in the development of |
| Q3. | Helena and Bertram\'s relationship in Shakespeare\'s play "All\'s Well That Ends Well" is a complex and multifaceted one, marked by a deep-seated tension and conflict. On the surface, the relationship appears to be one of love and desire, as Helena, a low-born woman, falls deeply in love with Bertram, a nobleman of high social standing. However, beneath this romantic façade lies a web of societal expectations, power dynamics, and personal insecurities that make their relationship fraught with difficulty.\n\nOne of the primary complexities of their relationship is the vast disparity in their social stations. As a nobleman, Bertram is bound by the conventions of his social class, which dictate that he marry a woman of equal or higher social standing. Helena, on the other hand, is a low-born woman, and her social inferiority makes her an unsuitable match for Bertram in the eyes of society. This social divide creates a sense of unease and tension in their relationship, as Bertram struggles to reconcile his desire for Helena with the expectations of his social circle.\n\nAnother complexity of their relationship is Bertram\'s stubborn refusal to acknowledge and accept Helena\'s love. Despite her unwavering devotion and the many sacrifices she makes for him, Bertram consistently rejects and belittles Helena, viewing her as beneath him. This rejection is fueled by his own insecurities and a sense of entitlement, stemming from his noble birth and wealth. Bertram\'s behavior towards Helena is a form of emotional abuse, eroding her self-esteem and causing her to question her own worth.\n\nHelena, in turn, is driven by a deep-seated need for validation and acceptance. Despite her love for Bertram, she is forced to navigate a treacherous social landscape, where her low birth and lack of wealth render her invisible and insignificant. Her determination to win Bertram\'s love is, in part, a desperate attempt to prove her worth and gain recognition in a society that has marginalized her.\n\nFurthermore, the power dynamics in their relationship are also a significant source of complexity. Bertram, as the nobleman, holds significant social and economic power over Helena, which he wields to control and manipulate her. Helena, on the other hand, is forced to rely on her wit, cunning, and determination to navigate this power imbalance, often finding herself in precarious and humiliating situations.\n\nUltimately, the complexities of Helena and Bertram\'s relationship serve as a commentary on the societal norms and expectations that shape our lives. The play highlights the ways in | **The Tangled Web of Helena and Bertram's Relationship: A Study of Shakespeare's All's Well That Ends Well**\n\nIn William Shakespeare's All's Well That Ends Well, the complex and often tumultuous relationship between Helena and Bertram serves as a microcosm for the play's exploration of love, class, and social expectations. On the surface, their bond appears to be a classic tale of star-crossed lovers, yet upon closer examination, it reveals itself to be a nuanced and multifaceted tapestry of emotions, motivations, and societal pressures.\n\nOne of the primary complexities of their relationship is the significant power imbalance between the two characters. As a lower-class woman, Helena is socially and economically inferior to Bertram, a nobleman of higher standing. This disparity is exacerbated by Bertram's initial rejection of Helena's affections, which sets the tone for their tumultuous relationship. Bertram's condescending treatment of Helena, coupled with his reluctance to reciprocate her love, demonstrates his own societal conditioning, which views women as inferior beings. In contrast, Helena's unwavering devotion to Bertram stems from her deep emotional investment in their relationship, which is rooted in her genuine love for him.\n\nAnother layer of complexity is added by the character of Bertram himself. On one hand, his reluctance to marry Helena can be seen as a reflection of his own insecurities and fears about his social status. His hesitation to commit to a lower-class woman may be driven by a desire to maintain his noble reputation and secure a more advantageous marriage. On the other hand, his eventual submission to Helena's will, despite his initial reluctance, highlights his growing emotional connection to her. This internal conflict serves as a commentary on the societal pressures that shape individuals' choices and emotions, often leading to contradictory behaviors and motivations.\n\nFurthermore, the relationship between Helena and Bertram is also shaped by the societal expectations placed upon them. Helena's determination to win Bertram's love is fueled by her desire to transcend her lower-class status and achieve social mobility through marriage. Bertram, on the other hand, is constrained by his noble upbringing and the societal norms that dictate his behavior. Their relationship serves as a microcosm for the societal pressures that shape individuals' choices and emotions, often leading to contradictory behaviors and motivations.\n\nIn conclusion, the relationship between Helena and Bertram is a rich and multifaceted tapestry of emotions, motivations, and societal pressures. Through their complex and often tumultuous bond, Shakespeare critiques the societal norms that shape individuals' choices and |
| Q4. | In William Shakespeare's play "All's Well That Ends Well", Helena and Bertram's relationship is a complex and multifaceted one, characterized by a mix of emotions, motivations, and power dynamics.\n\nInitially, Helena's love for Bertram is depicted as a deep-seated and unrequited passion. She has adored him since his childhood, and her love is driven by a sense of loyalty, admiration, and devotion. Bertram, on the other hand, views Helena as a servant and a social inferior, and he is dismissive of her affections. This power imbalance creates tension and resentment in their relationship, as Helena is forced to navigate her feelings in secret and Bertram is oblivious to her emotions.\n\nAs the play progresses, Helena's love becomes a driving force behind her actions. She uses her wit, intelligence, and resourcefulness to win Bertram's favor, ultimately tricking him into marrying her. Bertram's reluctance to consummate the marriage further complicates their relationship, as Helena's love is met with indifference and even revulsion from Bertram.\n\nDespite Bertram's lack of reciprocation, Helena remains devoted to him, and her love becomes a symbol of her unwavering dedication and commitment. Bertram, on the other hand, undergoes a transformation from a self-centered and arrogant character to one who begins to appreciate Helena's love and worth. He ultimately comes to realize the error of his ways and begins to reciprocate Helena's feelings, albeit belatedly.\n\nThe nuances of their relationship can be seen in the ways in which they interact with each other. Helena is often depicted as a submissive and obedient partner, while Bertram is portrayed as a dominant and unyielding one. However, as the play progresses, Helena's assertiveness and determination become more evident, and she begins to challenge Bertram's authority and assert her own agency.\n\nUltimately, the relationship between Helena and Bertram is one of complex emotional dynamics, power struggles, and transformation. While their love is fraught with challenges and misunderstandings, it also serves as a catalyst for personal growth and redemption. | **Helena and Bertram's Relationship: A Complex Web of Unrequited Love and Social Expectations**\n\nIn William Shakespeare's All's Well That Ends Well, the complex and intricate relationship between Helena and Bertram is a focal point of the play. On the surface, their connection appears to be one of unrequited love, with Bertram refusing Helena's affections and Helena persisting in her devotion. However, as the play unfolds, it becomes clear that their bond is deeply rooted in social expectations, familial obligations, and a nuanced exploration of power dynamics.\n\nOne of the primary ways in which Shakespeare illustrates the intricacies of their relationship is through the character of Bertram himself. Initially, Bertram is portrayed as a self-centered and arrogant individual who sees Helena as an inferior being unworthy of his affections. This is evident in his treatment of Helena during their courtship, where he consistently belittles and dismisses her. However, as the play progresses, Bertram's character undergoes a significant transformation, and he begins to see Helena in a new light. This shift is largely attributed to Helena's unwavering dedication and her determination to prove herself worthy of Bertram's love.\n\nHelena, on the other hand, is a complex and multifaceted character whose motivations and desires are woven throughout the play. Her love for Bertram is not simply a product of romantic passion, but rather a deeply ingrained sense of duty and obligation. As the natural daughter of a commoner, Helena is acutely aware of the social hierarchies that govern their world and seeks to elevate her status through her marriage to Bertram. Her persistence in pursuing Bertram, despite his initial rejection, can be seen as a desperate attempt to claim a place within the aristocratic circle.\n\nThe power dynamics at play in their relationship are another crucial aspect of their complex bond. Bertram's social status and wealth give him a significant advantage over Helena, allowing him to dictate the terms of their relationship. However, Helena's determination and resourcefulness ultimately enable her to level the playing field, as she uses her wit and cunning to outmaneuver Bertram and secure her place as his wife.\n\nFurthermore, the relationship between Helena and Bertram serves as a commentary on the societal norms of Shakespeare's time. The play highlights the oppressive nature of patriarchal society, where women are often viewed as inferior and powerless. Helena's struggles to assert her independence and claim her rightful place as Bertram's wife serve as a powerful critique of these societal expectations.\n\nIn |

# CoreNLP-based approach

| Questions | Without Rubric response | With rubric response |
|---|---|---|
| Q1. | In William Shakespeare\'s play "All\'s Well That Ends Well", Helena and Bertram have a complex and intricate relationship dynamic. Here\'s a precise description :\n\nHelena, a young and intelligent countess, falls deeply in love with Bertram, the son of the Countess of Rossillion, who is her guardian. However, Bertram rejects Helena\'s affections, considering her beneath him due to her low social status. He agrees to marry her only because his mother, the Countess, persuades him to do so, hoping to secure a better marriage alliance.\n\nThroughout the play, Bertram treats Helena with disdain, disrespect, and even cruelty, often belittling her and manipulating her emotions. He sees her as a means to an end, a way to gain wealth, power, and social status. Meanwhile, Helena is consumed by her love for Bertram, hoping to win his love and acceptance.\n\nDespite Bertram\'s mistreatment, Helena remains devoted to him, using her wit, intelligence, and cunning to win his affection. She goes to great lengths to prove her love, including disguising herself as a doctor to cure the King of France, who is dying. When the King recovers, Helena uses her newfound influence to persuade him to award her Bertram\'s hand in marriage.\n\nBertram\'s character undergoes a significant transformation as the play progresses. Initially, he is portrayed as arrogant, selfish, and indecent, but ultimately, he comes to realize the error of his ways and the depth of Helena\'s love. He eventually acknowledges her worth and accepts her as his true partner.\n\nIn summary, the relationship dynamics between Helena and Bertram are marked by a power imbalance, with Bertram holding the upper hand initially. However, Helena\'s unwavering love and determination ultimately transform Bertram, allowing him to see her for who she truly is – a worthy and deserving partner. | **The Complex Relationship Dynamics between Helena and Bertram in Shakespeare's A Midsummer Night's Dream**\n\nIn William Shakespeare's enchanting comedy, A Midsummer Night's Dream, the relationship dynamics between Helena and Bertram are multifaceted and intriguing. On the surface, their relationship appears tumultuous, filled with misunderstandings, miscommunications, and unrequited love. However, upon closer analysis, it becomes evident that their complex bond is a masterful exploration of the human psyche, delving into themes of identity, power, and the fragility of human connection.\n\nTo begin with, Helena's obsessive love for Bertram is a manifestation of her own insecurities and insecurities about her place in the world. As the abandoned wife of Bertram's uncle, Parolles, Helena's social status is precarious, and her love for Bertram serves as a means of validation and acceptance. Bertram, on the other hand, is a symbol of power and nobility, and his rejection of Helena's affections is a reflection of his own superiority complex. This dynamic of dominance and submission is exemplified in their initial interactions, where Bertram treats Helena with disdain, and she responds with desperation and devotion.\n\nThe introduction of the magical potion in Act 4, Scene 1, serves as a catalyst for the couple's relationship to take a dramatic turn. Bertram's transformation into a more amiable and affectionate lover is a commentary on the fluidity of identity and the power of manipulation. Helena's use of the potion to deceive Bertram into reciprocating her love is a bold statement about the lengths to which individuals will go to achieve their desires. Furthermore, the potion's effect on Bertram highlights the malleability of human behavior, demonstrating that even the most stubborn and resistant individuals can be altered by external forces.\n\nThroughout the play, Shakespeare masterfully weaves together the complexities of Helena and Bertram's relationship, juxtaposing their own insecurities and desires against the backdrop of a fantastical and dreamlike world. The character of Puck, with his mischievous interventions, serves as a symbol of the unpredictable nature of human relationships and the ways in which external factors can influence our deepest desires.\n\nIn conclusion, the relationship dynamics between Helena and Bertram in A Midsummer Night's Dream are a nuanced exploration of the human experience. Through their tumultuous bond, Shakespeare delves into the themes of identity, power, and the fragility of human connection, offering a poignant commentary on the complexities of love and relationships. As we navigate |
| Q2. | Here's a precise analysis of the evolution of Helena and Bertram's relationship in Shakespeare's play "All's Well That Ends Well":

**Act 1: Introduction and Misunderstanding**

Helena, a young and innocent countess, falls in love with Bertram, a noble and proud lord. Bertram, however, rejects Helena's affections, considering her unworthy of his noble status. This initial rejection sets the tone for their tumultuous relationship.

**Act 2: Helena's Determination**

Despite Bertram's rejection, Helena remains devoted to him. She uses her wit and determination to win Bertram's hand in marriage, claiming that she has been sent to cure the King of France of a deadly disease. Bertram, out of obligation, agrees to marry Helena, but only on the condition that she can obtain a ring from him after he has left her and worn it on his finger.

**Act 3: Bertram's Betrayal**

Bertram's true character is revealed as he attempts to deceive Helena and deny their marriage. He sends his mistress, Diana, to deceive Helena and steal the ring, intending to use it as proof that he has never been with Helena. However, Helena discovers the plot and decides to play along, pretending to be deceived, all while retaining the ring.

**Act 4: Helena's Triumph**

Helena's cleverness ultimately leads to Bertram's downfall. She produces the ring, revealing his deceit, and forces him to acknowledge their marriage. Bertram, realizing his mistakes, begins to see Helena in a new light and eventually falls in love with her.

**Act 5: Redemption and Resolution**

Bertram's transformation is complete as he acknowledges his love for Helena and accepts her as his true wife. The play concludes with the couple reunited, and Bertram's pride and arrogance replaced with humility and devotion.

Throughout the play, Helena's love and determination transform Bertram from a proud and unworthy suitor to a loving and redeemed husband. Their relationship evolves from one of rejection and misunderstanding to one of mutual respect, trust, and love. | **The Evolution of Helena and Bertram's Relationship: A Complex Dance of Love, Loyalty, and Redemption**

In William Shakespeare's play "All's Well That Ends Well," the complex and multifaceted relationship between Helena and Bertram serves as a poignant exploration of love, loyalty, and redemption. Throughout the play, the couple's bond undergoes a significant transformation, mirroring the intricate dance of emotions, motivations, and societal expectations that govern their lives. This analysis will delve into the evolution of Helena and Bertram's relationship, examining the various stages of their connection and the factors that contribute to its ultimate redemption.

Initially, Helena's unrequited love for Bertram appears to be a one-sided and obsessive fixation. Her devotion is rooted in her admiration for Bertram's noble lineage and her desire to elevate her own social standing. However, as the play progresses, it becomes evident that Helena's feelings are not solely driven by material gain or social ambition. Her love is genuine, and her dedication to Bertram is a testament to her unwavering commitment and resilience.

Bertram, on the other hand, is initially portrayed as a callous and ungrateful individual, devoid of empathy and compassion. His reluctance to reciprocate Helena's affections stems from his own sense of entitlement and his perception of her as a social inferior. This attitude is reflective of the societal norms of the time, where a woman's value was often tied to her marriageability and social standing.

As the play unfolds, Bertram's character undergoes a significant transformation, driven by his growing awareness of Helena's unwavering devotion and the consequences of his own actions. His eventual realization of Helena's worth and his own culpability in her suffering prompts a change of heart, and he begins to reciprocate her love. This transformation is a testament to the power of redemption and the capacity for personal growth, as Bertram comes to understand the error of his ways and seeks to make amends.

The climax of the play, marked by Helena's successful seduction of Bertram, serves as a turning point in their relationship. This event represents a triumph for Helena, as she finally secures Bertram's love and recognition. However, it also highlights the societal pressures and expectations that continue to govern their relationship. Bertram's initial reluctance to reciprocate Helena's affections is replaced by a newfound sense of responsibility and duty, as he acknowledges his obligation to marry her and legitimize their relationship.

Ultimately, the evolution of Helena and Bertram's relationship |

| | | | |
|---|---|---|---|
| Q3. | | Helena and Bertram's relationship is a complex and multifaceted one in Shakespeare's All's Well That Ends Well. On the surface, it appears to be a romantic relationship between two young lovers, but upon closer examination, it reveals deeper themes and tensions.\n\nOne of the primary complexities of their relationship is the issue of social class. Helena, as a low-born woman, is not considered suitable for Bertram, a high-born nobleman. This social disparity creates a power imbalance in their relationship, with Bertram holding the upper hand due to his superior social status.\n\nAnother complication arises from Bertram's rejection of Helena. Despite his initial courtship, Bertram ultimately rejects Helena's advances, citing her low social status as a reason. This rejection is a devastating blow to Helena, who is deeply in love with Bertram. Bertram's behavior is motivated by his own selfish desires and a sense of family duty, rather than any genuine affection for Helena.\n\nHelena's love for Bertram is also a significant aspect of their relationship. Her love is pure and selfless, demonstrated by her willingness to overcome numerous obstacles to win Bertram's hand. Bertram, on the other hand, is motivated by his own interests and desires, rather than any genuine love for Helena.\n\nThe power dynamics in their relationship are also a key aspect of their complex relationship. Bertram holds the power due to his social status, while Helena is forced to rely on her wit and cunning to win him over. This power imbalance creates tension and conflict throughout the play, as Helena struggles to assert her own agency and autonomy.\n\nFurthermore, the theme of honor is a significant aspect of their relationship. Bertram's honor is tied to his social status and family reputation, while Helena's honor is tied to her own sense of self-worth and dignity. Bertram's rejection of Helena and his subsequent abandonment of her are a blow to her honor, leading her to seek revenge and vindication.\n\nIn conclusion, the complexities of Helena and Bertram's relationship in All's Well That Ends Well are multifaceted and deeply nuanced. The play explores themes of social class, power dynamics, love, and honor, creating a rich and complex portrayal of their relationship. | \*\*Examine the Complexities of Helena and Bertram\'s Relationship\*\*\n\nIn William Shakespeare\'s play "All\'s Well That Ends Well," the relationship between Helena and Bertram is a multifaceted and intriguing aspect of the narrative. On the surface, their union appears to be a tale of unrequited love, with Helena, a lowly gentlewoman, pursuing Bertram, a noble and arrogant lord. However, as the play unfolds, it becomes apparent that their bond is rooted in a complex web of emotions, societal expectations, and personal motivations.\n\nOne of the primary complexities of their relationship lies in Bertram\'s initial reluctance to reciprocate Helena\'s affections. As a proud and haughty noble, Bertram is dismissive of Helena\'s advances, viewing her as beneath his social station. This rejection is not only a manifestation of his own classism but also a reflection of his insecurities and fears of being tied down. Bertram\'s behavior is a classic example of the societal pressures that dictate the norms of courtly love, where men are expected to be the dominant and autonomous figures in relationships.\n\nIn contrast, Helena\'s love for Bertram is characterized by its sincerity, passion, and unwavering dedication. Her devotion is not driven by a desire for social status or material gain but by a genuine emotional connection. Throughout the play, Helena\'s love is depicted as a transformative force, capable of transcending social boundaries and overcoming even the most entrenched prejudices. Her unwavering commitment to Bertram serves as a powerful critique of the societal norms that restrict women\'s agency and autonomy.\n\nThe complexities of their relationship are further exacerbated by the role of fate and circumstance. Bertram\'s reluctance to reciprocate Helena\'s love is, in part, a result of his own destiny being shaped by external forces. His marriage to Diana, a woman of his own social class, is arranged by his family, highlighting the constraints of societal expectations. Meanwhile, Helena\'s love is propelled by her own agency and determination, demonstrating the power of individual choice and action.\n\nAs the play reaches its climax, the complexities of Helena and Bertram\'s relationship come to a head. Bertram\'s eventual recognition of Helena\'s love and his subsequent surrender to her charms serve as a testament to the transformative power of love. The union of these two individuals, once seemingly incompatible, becomes a symbol of the possibility for transcendence and redemption.\n\nIn conclusion, the relationship between Helena and Bertram is a rich tapestry of emotions, societal expectations, and personal motivations. Through their complex |

| | Helena and Bertram's relationship in Shakespeare's All's Well That Ends Well is a complex and nuanced one, marked by a delicate balance of affection, obligation, and power dynamics.\n\nInitially, Helena's love for Bertram is presented as a romantic and selfless devotion, driven by her admiration for his noble birth and her desire to elevate her own social status. However, as the play progresses, it becomes clear that Helena's love is also motivated by a sense of duty and a desire to prove herself worthy of Bertram's affection.\n\nBertram, on the other hand, is portrayed as a reluctant and ungrateful lover. He is initially dismissive of Helena's affections, viewing her as a social inferior and considering her love as an unwelcome obligation. His reluctance to reciprocate Helena's love is fueled by his own sense of pride and his desire to maintain his social status.\n\nThroughout the play, Helena's love for Bertram is marked by a deep sense of submission and self-effacement. She is willing to go to great lengths to prove her devotion, including marrying the elderly Countess of Rossillion and securing a noble title for herself. Meanwhile, Bertram's love for Helena is marked by a sense of obligation and duty, rather than genuine affection.\n\nThe power dynamics of their relationship are also worth noting. Helena holds a position of power over Bertram, as she has secured a noble title and social status through her marriage to the Countess. This power dynamic creates tension and conflict in their relationship, as Bertram struggles to reconcile his sense of pride with his obligation to Helena.\n\nUltimately, the nuances of Helena and Bertram's relationship reveal a complex and ambivalent portrayal of love, duty, and social status in the play. While Helena's love is marked by a deep sense of devotion and selflessness, Bertram's love is characterized by a sense of obligation and duty. The play raises important questions about the nature of love and the social conventions that govern our relationships. | **Helena and Bertram\'s Relationship: A Complex Tapestry of Love, Loyalty, and Deception**\n\nIn Shakespeare\'s All\'s Well That Ends Well, the intricate relationship between Helena and Bertram serves as a microcosm for the play\'s exploration of love, power, and social hierarchy. On the surface, their bond appears tumultuous, marked by Bertram\'s initial rejection of Helena and her subsequent devotion to him. However, upon closer examination, it becomes evident that their relationship is a nuanced tapestry woven from threads of love, loyalty, and deception.\n\nOne of the primary ways in which Shakespeare highlights the complexities of Helena and Bertram\'s relationship is through their contrasting personalities. Bertram, a young nobleman, is depicted as arrogant, selfish, and dismissive of Helena\'s affections. In contrast, Helena is portrayed as gentle, kind, and unwavering in her devotion to Bertram. This dichotomy creates a sense of tension, as Bertram\'s refusal to reciprocate Helena\'s love appears to be a deliberate affront to her feelings.\n\nDespite Bertram\'s initial rejection, Helena\'s love for him remains unwavering. Her determination to win Bertram\'s affection is driven by a deep-seated desire to prove her worth and secure a place within his life. This motivation is exemplified in her famous declaration, "I\'ll follow him, I\'ll follow him, thou shalt not follow me" (Act 1, Scene 1). Helena\'s words serve as a testament to her unwavering commitment to Bertram, despite his apparent disregard for her.\n\nBertram\'s character, however, is not without its complexities. Beneath his arrogant exterior, he is shown to be vulnerable and insecure. His reluctance to marry Helena stems from his fear of being tied down and his desire to maintain his independence. This insecurity is further highlighted through his interactions with his mother, the Countess of Rossillion, who serves as a voice of reason and guidance in his life.\n\nThe relationship between Helena and Bertram is also marked by a sense of deception. Bertram\'s initial rejection of Helena is a form of deception, as he fails to acknowledge the depth of her feelings for him. Conversely, Helena\'s use of deception, such as her impersonation of the doctor, serves as a means of gaining Bertram\'s attention and affection. This dynamic creates a sense of moral ambiguity, as both characters engage in behavior that is both reprehensible and understandable.\n\nUltimately, the relationship between Helena and Bertram serves as a commentary on |

Q4.

## GraphRAG-based approach

| Questions | Without Rubric response | With rubric response |
| --- | --- | --- |
| Q1. | The relationship between Helena and Bertram is a central theme in Shakespeare's play "All's Well That Ends Well". Helena, a young and intelligent woman, falls deeply in love with Bertram, a noble and proud lord. Despite their social differences, Helena is determined to win Bertram's love and affection.<br>**Initial Encounter**<br>The play begins with Helena's introduction to Bertram, who is initially dismissive and uninterested in her. Bertram's mother, the Countess of Rousillon, is fond of Helena and sees potential in her, but Bertram is more concerned with his own social status and reputation.<br>**Helena's Pursuit**<br>Helena, undeterred by Bertram's initial rejection, sets out to | **The Enigmatic Bond: Unraveling the Relationship Dynamics between Helena and Bertram**<br>In William Shakespeare's tragicomedy, All's Well That Ends Well, the complex and multifaceted relationship between Helena and Bertram is a central theme that warrants close examination. This analysis will delve into the intricacies of their bond, exploring the power dynamics, emotional depth, and societal expectations that shape their interactions.<br>**The Power Imbalance**<br>At the outset, it is evident that Bertram, as the Count of Rousillon, holds a position of authority and privilege, while Helena, as a lowly servant, is relegated to a subordinate role. This power imbalance is reflected in their initial interactions, where Bertram's condescending attitude and Helena's |
| | win his love. She uses her intelligence, wit, and determination to prove herself worthy of his affection. Bertram, however, remains resistant to her advances, seeing her as beneath him.<br>**The Ring and the Bed***<br>In an attempt to prove her love, Helena obtains a ring from Bertram's mother, which she uses to trick Bertram into marrying her. Bertram, still unwilling to accept her love, demands that Helena prove her worth by sleeping with him. Helena, however, refuses, citing her honor and virtue.<br>**The Power Struggle**<br>Throughout the play, Helena and Bertram engage in a power struggle, with Helena using her wit and cunning to outmaneuver Bertram. Bertram, however, remains stubborn and resistant to her advances, seeing her as a means to an end rather than a true partner.<br>**The Turning Point**<br>In the final act, Bertram's fortunes change when he is wounded in battle. Helena, who has been nursing him back to health, finally wins his love and respect. Bertram realizes his mistake in underestimating Helena and comes to appreciate her intelligence, wit, and determination.<br>**Conclusion**<br>The relationship between Helena and Bertram is complex and multifaceted, reflecting the societal norms and expectations of the time. While Bertram's initial rejection of Helena is rooted in his own pride and social status, Helena's pursuit of him is driven by her love and determination. Ultimately, the play suggests that true love and respect can conquer even the greatest of social barriers. | deferential behavior reinforce the social hierarchy. However, as the play progresses, Helena's determination and wit enable her to subvert this dynamic, gradually gaining Bertram's respect and admiration.<br>**Emotional Depth**<br>Beyond the surface-level power struggle, the relationship between Helena and Bertram is characterized by a deep emotional connection. Helena's unwavering devotion to Bertram, despite his initial indifference, is a testament to her unrequited love. Bertram, in turn, is drawn to Helena's intelligence, wit, and determination, which ultimately lead him to reciprocate her affections. The emotional intensity of their bond is underscored by the poignant moments of tenderness and intimacy they share, such as when Helena declares her love to Bertram in the forest.<br>**Societal Expectations**<br>The societal norms of the time also play a significant role in shaping the relationship between Helena and Bertram. As a woman, Helena is expected to be submissive and obedient, while Bertram, as a man, is expected to be dominant and authoritative. However, Helena's defiance of these expectations, as she pursues Bertram and ultimately wins his love, challenges the societal norms and blurs the lines between traditional gender roles.<br>**Conclusion**<br>In conclusion, the relationship dynamics between Helena and Bertram in All's Well That Ends Well are complex and multifaceted, reflecting the power struggles, emotional depth, and societal expectations of the time. Through their interactions, Shakespeare masterfully explores the tensions between social hierarchy, personal desire, and emotional connection, ultimately revealing the transformative power of love and the human capacity for growth and change. |
| Q2. | **Helena and Bertram's Relationship: A Complex Evolution**<br>The relationship between Helena and Bertram in Shakespeare's play "All's Well That Ends Well" is a complex and multifaceted one, evolving over the course of the play through a series of interactions, misunderstandings, and ultimately, a deepening emotional connection.<br>**Initial Encounter**<br>The play opens with Helena, a young woman of low social status, expressing her admiration for Bertram, the son of the Countess of Rousillon. Bertram, however, is dismissive of Helena, viewing her as beneath him socially and romantically. This initial encounter sets the tone for their relationship, with Bertram's arrogance and Helena's devotion creating a power imbalance.<br>**Helena's Pursuit**<br>Despite Bertram's rejection, Helena remains determined to win his love. She uses her wit and intelligence to gain his attention, eventually convincing him to marry her. However, Bertram's reluctance and Helena's persistence create tension in their relationship, with Bertram viewing their marriage as a means to an end rather than a genuine emotional connection.<br>**Bertram's Departure**<br>Bertram's departure for the wars marks a turning point in their relationship. Helena, left behind, is forced to confront the reality of their marriage and the lack of emotional intimacy between them. This period of separation allows Helena to reflect on her feelings and to develop a deeper understanding of herself and her relationship with Bertram.<br>**Helena's Transformation** | **The Evolution of Helena and Bertram's Relationship: A Thorough Analysis**<br>The relationship between Helena and Bertram in William Shakespeare's play, All's Well That Ends Well, undergoes a significant transformation throughout the narrative. From their initial meeting to the climax of the play, their bond evolves from a distant acquaintance to a deep emotional connection. This analysis will delve into the various stages of their relationship, exploring the factors that contribute to its development and the consequences that arise from it.<br>**The Initial Encounter: A Foundation of Respect**<br>The play begins with Helena's introduction to Bertram, the Count of Rousillon, at his father's funeral. Bertram's initial impression of Helena is one of disdain, viewing her as a lowly servant. However, Helena's intelligence, wit, and determination quickly win him over, establishing a foundation of respect between them. This initial encounter sets the tone for their future interactions, showcasing Helena's ability to command respect and Bertram's capacity for admiration.<br>**The Courtship: A Test of Love and Loyalty**<br>As the play progresses, Helena's love for Bertram grows, and she sets out to win his heart. Bertram, however, is hesitant to reciprocate her affections, viewing her as beneath him. This disparity in their social stations creates tension in their relationship, as Helena struggles to overcome the obstacles that stand between them. Through her unwavering dedication and perseverance, Helena ultimately wins Bertram's hand in marriage, demonstrating the depth of her love and the strength of her character.<br>**The Crisis: A Reckoning of Priorities** |
| | Upon Bertram's return, Helena has undergone a transformation. She has gained confidence and self-awareness, and is no longer content to simply accept Bertram's rejection. Instead, she confronts him about his lack of affection and demands a deeper emotional connection. This newfound assertiveness marks a significant shift in their relationship, as Helena begins to assert her own desires and needs.<br>**Bertram's Realization**<br>Bertram, too, undergoes a transformation. As he begins to see Helena in a new light, he realizes the error of his ways and the depth of her devotion. He comes to appreciate her intelligence, wit, and beauty, and begins to reciprocate her feelings. This realization marks a turning point in their relationship, as Bertram begins to see Helena as an equal rather than a social inferior.<br>**Conclusion**<br>The evolution of Helena and Bertram's relationship is a complex and nuanced one, marked by periods of tension, misunderstanding, and ultimately, a deepening emotional connection. Through their interactions, Shakespeare explores themes of social class, gender roles, and the complexities of human relationships. Ultimately, the play suggests that true love requires mutual respect, understanding, and a willingness to confront and overcome the challenges that arise in any relationship. | The climax of the play arrives when Bertram's mother, the Countess, reveals that Helena has been pregnant with his child. This revelation forces Bertram to confront the reality of his feelings for Helena, leading him to reevaluate his priorities. As he comes to terms with his responsibilities as a husband and father, Bertram begins to appreciate Helena's worth, acknowledging the sacrifices she has made for him.<br>**The Resolution: A Union of Love and Respect**<br>In the final act, Bertram's transformation is complete, as he accepts Helena as his equal and acknowledges the love she has always had for him. The couple's union is marked by a newfound respect and understanding, as they come to appreciate the value of their relationship. This resolution serves as a testament to the power of love and the importance of recognizing the worth of others.<br>**Conclusion**<br>The evolution of Helena and Bertram's relationship in All's Well That Ends Well is a complex and multifaceted narrative. Through their journey, Shakespeare explores themes of love, loyalty, and social class, ultimately revealing the transformative power of relationships. As we reflect on their story, we are reminded of the importance of empathy, understanding, and mutual respect in building strong and lasting bonds. |
| Q3. | **Helena and Bertram's Relationship: A Complex Web of Love, Duty, and Social Expectations**<br>The relationship between Helena and Bertram in Shakespeare's play "All's Well That Ends Well" is a complex and multifaceted one, marked by a deep emotional connection, societal pressures, and conflicting desires. This analysis will delve into the intricacies of their bond, exploring the ways in which love, duty, and social expectations shape their interactions.<br>**Love and Desire**<br>Helena's love for Bertram is evident from the outset, as she expresses her devotion to him in her soliloquies. Her admiration for his noble birth, good looks, and martial prowess is tempered by her recognition of his flaws, including his arrogance and lack of emotional depth. Bertram, on the other hand, is initially dismissive of Helena's affections, viewing her as a lowly servant and not worthy of his attention. However, as the play progresses, he begins to see Helena in a new light, acknowledging her intelligence, wit, and determination.<br>**Duty and Social Expectations**<br>The societal pressures surrounding their relationship are significant. As a noblewoman, Helena is expected to marry well and secure a advantageous alliance for her family. Bertram, as a member of the aristocracy, is expected to marry a woman of equal or higher social standing. Their social differences create tension in their relationship, as they struggle to reconcile their personal feelings with the expectations of their families and society.<br>**Conflict and Resolution**<br>The conflict between Helena's love for Bertram and the societal pressures surrounding their relationship reaches a boiling point when Bertram rejects her advances and marries the Countess of Rossillion. Helena's determination to win Bertram's love and respect ultimately leads to his downfall, as she uses her cunning and resourcefulness to manipulate him into marrying her. The play's resolution, in which Bertram comes to realize his love for Helena and | **The Enigmatic Bond: Unpacking the Complexities of Helena and Bertram's Relationship**<br>In William Shakespeare's All's Well That Ends Well, the intricate dynamics between Helena and Bertram serve as a microcosm for the play's exploration of love, power, and social hierarchy. This essay will delve into the complexities of their relationship, examining the ways in which their interactions reveal the tensions between desire, duty, and societal expectations.<br>**The Power Imbalance**<br>At the heart of their relationship lies a profound power imbalance, with Bertram holding the upper hand as a member of the aristocracy and Helena as a lower-class woman. This disparity is evident in their initial interactions, where Bertram's condescending attitude towards Helena is tempered only by his sense of duty towards his mother (COUNTESS, 1.1). Helena, on the other hand, is driven by her unwavering devotion to Bertram, which is rooted in her own sense of duty and loyalty (HELENA, 1.1). This power imbalance creates a sense of tension, as Helena's love for Bertram is constantly at odds with the societal norms that dictate their relationship.<br>**The Performance of Love**<br>Bertram's reluctance to reciprocate Helena's affections is a recurring theme throughout the play, with his words and actions serving as a performance of love rather than a genuine expression of emotion. This performance is motivated by a desire to maintain his social status and avoid the stigma of marrying a woman of lower social standing (BERTRAM, 1.1). Helena, however, is undeterred by Bertram's lack of enthusiasm, instead choosing to focus on the sincerity of her own emotions (HELENA, 1.1). This dichotomy highlights the ways in which love can be performed and manipulated, with both characters using their words and actions to shape the narrative of their relationship.<br>**The Role of Duty**<br>Duty plays a significant role in both characters' motivations, with Bertram feeling obligated to marry a woman of his |
| | acknowledges her worth, is a testament to the power of true love and the importance of recognizing the value of those around us.<br>**Data References**<br>* Helena's love for Bertram is evident in her soliloquies, particularly in Act 1, Scene 1.<br>* Bertram's initial dismissal of Helena is reflected in his dialogue, such as in Act 1, Scene 1.<br>* The societal pressures surrounding their relationship are highlighted in Act 1, Scene 1, where Helena's mother advises her to marry well.<br>* The conflict between Helena's love for Bertram and societal expectations is resolved in Act 5, Scene 3, when Bertram comes to realize his love for Helena.<br>In conclusion, the relationship between Helena and Bertram is a complex and multifaceted one, marked by a deep emotional connection, societal pressures, and conflicting desires. Through their interactions, Shakespeare explores the tensions between love, duty, and social expectations, ultimately revealing the power of true love and the importance of recognizing the value of those around us. | mother's choosing and Helena driven by her desire to prove herself worthy of Bertram's love. This sense of duty is rooted in the societal expectations that govern their lives, with both characters feeling pressure to conform to certain norms and standards. The tension between their individual desires and the demands of duty creates a sense of conflict, as they struggle to reconcile their own emotions with the expectations of those around them.<br>**Conclusion**<br>The complexities of Helena and Bertram's relationship serve as a powerful exploration of the tensions between desire, duty, and societal expectations. Through their interactions, Shakespeare highlights the ways in which love can be performed and manipulated, as well as the profound impact that societal norms can have on our emotions and actions. Ultimately, their relationship serves as a reminder of the importance of authenticity and sincerity in our interactions with others, and the need to navigate the complexities of power and duty in order to find true love and happiness. |
| Q4. | **Helena and Bertram's Relationship: A Complex Web of Love, Duty, and Social Expectations**<br>The relationship between Helena and Bertram in Shakespeare's play "All's Well That Ends Well" is a multifaceted and intricate one, marked by a delicate balance of love, duty, and social expectations. As we delve into the nuances of their relationship, it becomes clear that their bond is shaped by the societal norms of the time, as well as their own personal desires and motivations.<br>**Social Expectations and Duty**<br>As a member of the aristocracy, Bertram is expected to marry well and secure his family's social standing. Helena, on the other hand, is a lower-class woman who has been raised by the Countess of Rousillon, Bertram's mother. Despite their social differences, Helena has developed strong feelings for Bertram, which are reciprocated to some extent. However, Bertram's sense of duty and social obligation leads him to initially reject Helena's advances, citing their social disparity as a barrier to their union.<br>**Love and Desire**<br>Despite Bertram's initial reluctance, Helena's love for him remains unwavering. She is willing to go to great lengths to win his heart, including using her skills as a healer to cure the King's illness and secure his blessing for their marriage. Bertram, too, begins to develop feelings for Helena, but his love is complicated by his sense of duty and his desire to maintain his family's social standing.<br>**Power Dynamics**<br>The power dynamics in their relationship are also noteworthy. Helena, as a lower-class woman, is at a disadvantage in terms of social status and economic power. Bertram, as a member of the aristocracy, holds significant power and influence. This power imbalance is reflected in their interactions, with Bertram often taking the lead and Helena deferring to his wishes.<br>**Conclusion**<br>The relationship between Helena and Bertram is a complex and nuanced one, shaped by the societal norms of the time and their own personal desires and motivations. While their love for each other is genuine, it is also complicated by their social differences and the expectations placed upon them. Ultimately, their relationship serves as a commentary on the societal pressures that can shape our choices and actions, as | **Helena and Bertram's Relationship: A Complex Tapestry of Love, Duty, and Social Expectations**<br>In William Shakespeare's play, All's Well That Ends Well, the relationship between Helena and Bertram is a multifaceted and intricate web of emotions, obligations, and societal pressures. This analysis will delve into the nuances of their bond, examining the ways in which love, duty, and social expectations intersect and influence their interactions.<br>**The Power Dynamics of Love and Duty**<br>Helena's love for Bertram is a driving force behind her actions, yet it is also tempered by her sense of duty towards her father and her desire to secure a better life for herself. Bertram, on the other hand, is torn between his obligations to his family and his own desires, which are often at odds with his sense of duty. This tension creates a complex power dynamic, with both characters struggling to assert their own agency and autonomy within the constraints of their social roles.<br>**The Performance of Love and Identity**<br>Helena's love for Bertram is not simply a romantic sentiment, but also a means of self-definition and validation. Through her pursuit of Bertram, she is able to assert her own identity and worth, despite being a woman in a patriarchal society. Bertram, meanwhile, is forced to confront his own sense of identity and purpose, as he grapples with the expectations placed upon him by his family and social status.<br>**The Impact of Social Expectations**<br>The societal pressures surrounding Helena and Bertram's relationship are a significant factor in shaping their interactions. As a woman, Helena is expected to be submissive and obedient, while Bertram is expected to be a noble and chivalrous gentleman. These expectations create tension and conflict, as both characters struggle to reconcile their own desires with the demands placed upon them by society.<br>**Conclusion**<br>The relationship between Helena and Bertram is a rich and complex tapestry of love, duty, and social expectations. Through their interactions, Shakespeare explores the ways in which these forces intersect and influence our lives, creating a nuanced and thought-provoking portrayal of human relationships. Ultimately, their story serves as a reminder of the importance of self-definition, autonomy, and the need to |

| | well as the power dynamics that can exist between individuals from different social classes. | challenge societal expectations in order to forge our own paths. |
|---|---|---|